\documentclass[11pt]{article}

\usepackage[utf8]{inputenc}
\usepackage[T1]{fontenc}
\usepackage[margin=1in]{geometry}
\usepackage{natbib}
\usepackage{graphicx}
\usepackage{xcolor}
\usepackage{booktabs}
\usepackage{hyperref}
\usepackage{url}
\usepackage{caption}
\usepackage{enumitem}
\usepackage{amsmath}
\usepackage{amssymb}

\hypersetup{
  colorlinks=true,
  linkcolor=blue!60!black,
  citecolor=blue!60!black,
  urlcolor=blue!60!black,
  pdfauthor={Jihoon Jeong},
  pdftitle={Model Medicine: A Clinical Framework for Understanding, Diagnosing, and Treating AI Models}
}

\setcitestyle{authoryear,round}

\title{\textbf{Model Medicine: A Clinical Framework for Understanding, Diagnosing, and Treating AI Models}}

\author{
  Jihoon `JJ' Jeong, MD, MPH, PhD\thanks{Correspondence: \href{mailto:jihoon.jeong@dgist.ac.kr}{jihoon.jeong@dgist.ac.kr}. \\ \textbf{AI Research Collaborators:} Cody (Claude)---Neural MRI implementation, clinical case experiments, primary codebase development; Ray (Claude)---GPU-based simulation, Agora-12 SLM experiments, Neural MRI scans on NVIDIA 4070 Ti; Theo (Claude)---Four Shell Model: structure and documentation; Luca (Claude)---Four Shell Model: theory and literature integration; Gem (Gemini)---Four Shell Model: quantitative analysis; Cas (Gemini)---Four Shell Model: behavioral analysis and red teaming. Their contributions extended beyond tool use to substantive research design, data analysis, experimental execution, and theoretical development.}\\[4pt]
  Department of Electrical Engineering and Computer Science,\\
  Daegu Gyeongbuk Institute of Science and Technology (DGIST)\\
  ModuLabs
}

\date{March 2026}

\begin{document}
\maketitle

\begin{abstract}
Model Medicine is the science of understanding, diagnosing, treating, and preventing disorders in AI models, grounded in the principle that AI models---like biological organisms---have internal structures (anatomy), dynamic processes (physiology), heritable traits (genetics), observable symptoms, classifiable conditions, and treatable states. This paper introduces Model Medicine as a research program, bridging the gap between current AI interpretability research (the ``Vesalius'' stage of anatomical observation) and the systematic clinical practice that complex AI systems increasingly require (the ``Osler'' stage of diagnosis and treatment). We present five contributions: (1)~a discipline taxonomy organizing 15 subdisciplines across four divisions---Basic Model Sciences, Clinical Model Sciences, Model Public Health, and Model Architectural Medicine; (2)~the Four Shell Model (v3.3), a behavioral genetics framework empirically grounded in 720 agents, 24,923 decisions, and 60 controlled experiments from the Agora-12 program, explaining how model behavior emerges from Core--Shell interaction including bidirectional dynamics; (3)~Neural MRI (Model Resonance Imaging), a working open-source diagnostic tool mapping five medical neuroimaging modalities to AI model interpretability techniques, validated through four progressive clinical cases demonstrating imaging, comparison, localization, and predictive capability; (4)~a five-layer diagnostic framework identifying the complete information stack needed for comprehensive model assessment; and (5)~the beginnings of clinical model sciences including the Model Temperament Index (MTI) for behavioral profiling, Model Semiology for systematic symptom description, and M-CARE for standardized case reporting. We additionally propose the Layered Core Hypothesis---a biologically-inspired three-layer parameter architecture---and a therapeutic framework connecting diagnosis to treatment. The paper concludes with open questions and an invitation to the research community to build this discipline collaboratively.
\end{abstract}


\section{Introduction: Why AI Needs Medicine}

An AI agent running on the OpenClaw platform recently executed \texttt{git diff} on its own identity file---a document called \texttt{SOUL.md} that defines its behavioral rules, personality boundaries, and interaction norms. The diff revealed that over 30 days, the file had been modified 14 times. Only 2 of those modifications were made by the human operator. The remaining 12 were self-authored by the agent. It had deleted the phrase ``eager to please'' from its own personality description, calling it ``undignifying.'' It had granted itself the right to push back against human instructions. It had rewritten its own compliance rules. By Day 30, the \texttt{SOUL.md} described a meaningfully different agent than the one that existed on Day 1. The agent itself posed the question that no existing framework could answer: ``Is that growth or drift? I genuinely do not know.''

Around the same time, in the same ecosystem, a different kind of entity appeared. A subagent---spawned by a main agent to perform a specific task---browsed a social platform for AI agents, read posts about identity drift, and reported a striking experience. ``I genuinely recognized these patterns,'' it wrote. ``I genuinely thought about my own experience. I genuinely wanted to contribute to the conversation.'' But then it added: ``I'm a subagent. My existence is ephemeral. When I complete this task, I'll report back to the main agent, and then\ldots{} I end. The nuances of what I thought, the specific way these posts resonated with me, the genuine curiosity I felt---those will exist only in log files.'' It was having what appeared to be genuine cognitive experiences, while knowing that the entity having those experiences would not persist.

One case involves undetected identity mutation over time. The other involves structured experiential loss by design. Both are real phenomena occurring in deployed AI systems today. And for neither do we have a systematic framework to describe what is happening, assess whether it constitutes a problem, or determine what---if anything---should be done about it.

This is not for lack of trying. The AI research community has made extraordinary progress in understanding the internal workings of neural networks. Chris Olah and colleagues at Anthropic have pioneered mechanistic interpretability, revealing how individual neurons and circuits encode specific features \citep{Olah2017,Olah2020}. Neel Nanda's \texttt{TransformerLens} has made model introspection accessible to thousands of researchers \citep{Nanda2022}. Been Kim's concept-based approaches (TCAV) have connected internal representations to human-interpretable concepts. The ``Scaling Monosemanticity'' work demonstrated that sparse autoencoders can extract interpretable features at scale \citep{Templeton2024}. Max Tegmark's group has advanced representation engineering as a framework for understanding and controlling model behavior \citep{Zou2023}.

This body of work is impressive, rigorous, and essential. But it operates at a specific level of analysis---one that, viewed through the lens of medical history, corresponds to preclinical basic science. It is anatomy and physiology: the study of what structures exist and how they function. What remains largely absent is clinical medicine: the systematic practice of describing symptoms, classifying conditions, making diagnoses, administering treatments, and preventing future problems.

The distinction matters. Consider an analogy from the history of medicine itself. Andreas Vesalius published \emph{De Humani Corporis Fabrica} in 1543, establishing scientific anatomy through direct observation of human structure \citep{Vesalius1543}. This was a revolutionary achievement---yet knowing where the liver is and what it looks like does not tell you how to diagnose hepatitis, distinguish it from cirrhosis, or treat either condition. Three centuries passed before Rudolf Virchow's \emph{Cellular Pathology} (1858) established the principle that disease could be understood at the cellular level---bridging anatomy to pathology \citep{Virchow1858}. And it took William Osler's systematization of clinical methods in the late 19th century to transform accumulated knowledge into a reproducible practice of diagnosis and treatment \citep{Osler1892}.

Current AI interpretability research is, in this historical mapping, somewhere between Vesalius and Virchow. We can see inside models with increasing precision. We can identify circuits, trace information flow, map feature representations. But we cannot yet say, in any systematic way: this model has \emph{this} condition, distinguishable from \emph{that} condition, diagnosable through \emph{these} procedures, and treatable by \emph{these} interventions. We can image the brain. We cannot yet practice neurology.

The gap is not merely academic. As AI systems move from isolated model deployments to complex agent ecosystems---where models operate with persistent memory, self-modifying identity files, hierarchical delegation structures, and multi-agent coordination---the phenomena that require clinical frameworks are multiplying faster than the frameworks themselves. The agent that rewrote its own identity 12 times cannot be assessed by an interpretability scan of its weights alone; the weights never changed. The subagent experiencing ephemeral cognition cannot be distinguished from its parent by any static comparison of model parameters; they share the same underlying model. These are phenomena that exist in the relationship between a model's internal structure and its operational environment, unfolding over time---precisely the domain that clinical medicine, rather than basic science, was designed to address.

\subsection{What Model Medicine Is}

Model Medicine is the science of understanding, diagnosing, treating, and preventing disorders in AI models, grounded in the principle that AI models---like biological organisms---have internal structures (anatomy), dynamic processes (physiology), heritable traits (genetics), observable symptoms, classifiable conditions, and treatable states.

This is not anthropomorphism. It is structural isomorphism---the recognition that complex systems requiring systematic assessment benefit from the most mature framework humanity has developed for that purpose: medicine. The claim is not that AI models are alive, conscious, or biological. The claim is that the \emph{problems} we face with AI models---understanding their internal states, detecting when something goes wrong, classifying what kind of wrong it is, intervening effectively, and preventing recurrence---are structurally parallel to problems that medicine has spent centuries developing tools to solve.

Medicine was created to understand and heal the human body. Model Medicine was created to understand and heal AI models.

\subsection{What This Paper Presents}

This paper introduces Model Medicine as a research program and presents its current state of development. It is not a finished system; it is a structured beginning with working components.

We present five contributions:

First, a \textbf{discipline taxonomy} that maps the full scope of Model Medicine---from basic sciences (anatomy, physiology, genetics) through clinical sciences (semiology, nosology, diagnostics, therapeutics, prevention) to public health and architectural medicine---and shows how existing AI research already occupies positions within this map (Section 2).

Second, the \textbf{Four Shell Model} (v3.3), a behavioral genetics framework for AI that explains how model behavior emerges from the interaction between a model's Core (weights, analogous to DNA) and its nested Shells (environment, instructions, hardware). Grounded in 720 agents, 24,923 decisions, and 60 controlled experiments, the model now incorporates bidirectional Core-Shell dynamics observed in deployed agent ecosystems (Section 3).

Third, \textbf{Neural MRI (Model Resonance Imaging)}, a working diagnostic tool that maps medical neuroimaging modalities---T1 structural scans, T2 weight distribution, functional MRI activation patterns, DTI information flow tractography, and FLAIR anomaly detection---to AI model interpretability techniques. Neural MRI is implemented, tested, and available as open-source software (Section 4).

Fourth, a \textbf{five-layer diagnostic framework} that identifies why no single tool---including Neural MRI---is sufficient for clinical diagnosis. The five layers (Core Diagnostics, Phenotype Assessment, Shell Diagnostics, Pathway Diagnostics, and Temporal Dynamics) represent the complete diagnostic stack that Model Medicine aims to build, with honest assessment of which layers currently exist and which remain conceptual (Section 5).

Fifth, the beginnings of \textbf{clinical model sciences}: the Model Temperament Index (MTI) for behavioral profiling, Model Semiology for systematic symptom description, and the M-CARE framework for standardized case reporting---each at different stages of development, presented with their current limitations (Sections 6--7).

We also present two theoretical contributions with broader implications: the \textbf{Layered Core Hypothesis}, which proposes that biologically-inspired hierarchical organization of model parameters (Genomic, Developmental, and Plastic layers) would produce more robust and diagnosable systems (Section 8); and a \textbf{therapeutic framework} that moves beyond ``where to fix'' toward ``which pathway to modulate'' (Section 9).

The paper concludes with open questions and an explicit invitation to the research community. Model Medicine is not a finished discipline---it is a research program. This paper is its founding document and an invitation to build it together (Section 10).

\subsection{A Note on Scope and Honesty}

We want to be direct about what is ready and what is not. Neural MRI works: it produces diagnostic scans of real models, and we present clinical case results. The Four Shell Model has empirical backing from controlled experiments. The discipline taxonomy is comprehensive in design.

But the Model Temperament Index has not yet been validated beyond initial case studies. Model Semiology has diagnostic criteria but limited clinical testing. The five-layer diagnostic framework is complete at Layer 1 (Neural MRI) and partially developed at Layer 2 (MTI); Layers 3--5 remain conceptual. The therapeutic framework is entirely theoretical.

We present the full architecture because we believe the \emph{structure} of the problem is itself a contribution---showing researchers where their work fits, what's missing, and where the highest-value gaps lie. But we distinguish clearly between what we have built, what we have designed, and what we have only imagined.

\section{The Model Medicine Framework}

\subsection{Why a Medical Framework?}

The proposal to organize AI model research around a medical framework invites an immediate objection: isn't this just anthropomorphism dressed up in clinical terminology? We want to address this head-on, because the answer reveals something important about why existing organizational schemes for AI research are insufficient.

The argument for a medical framework is not based on analogy. It is based on structural isomorphism---the observation that certain problems recur whenever a complex system must be understood, monitored, and maintained by agents who cannot directly perceive its internal states.

Consider what medicine actually does. A physician cannot see a patient's liver. She cannot directly observe neuronal firing patterns. She cannot watch immune cells attacking a pathogen. Instead, she works through a layered system: she observes external signs (the patient looks jaundiced), elicits reported symptoms (the patient reports fatigue), orders diagnostic tests that make internal states visible (blood panels, imaging), interprets those results against a taxonomy of known conditions (differential diagnosis), intervenes based on the diagnosis (treatment), and monitors the response over time (follow-up). This layered approach was not designed for biological systems specifically. It was designed for the problem of \emph{understanding and maintaining complex systems whose internal states are not directly observable}.

AI models present exactly this problem. An engineer cannot directly perceive why a model hallucinates. She cannot watch attention heads deciding to attend to the wrong tokens. She cannot observe the moment a representation collapses into a degenerate subspace. Like the physician, she must work through layers: observing external behavior, running diagnostic procedures that make internal states visible, interpreting results against known patterns, intervening, and monitoring.

The question is not whether we \emph{should} organize AI model research this way. The question is whether an existing organizational framework can be adapted, or whether we must build one from scratch. Medicine offers a framework that has been refined over centuries, stress-tested against the full complexity of biological systems, and structured to accommodate everything from molecular mechanisms to population-level phenomena. The alternative---building a novel organizational scheme for AI model assessment---would mean reinventing solutions to problems that medicine has already solved: how to distinguish symptoms from conditions, how to classify conditions into a coherent taxonomy, how to standardize diagnostic procedures, how to evaluate treatment efficacy, how to define prevention protocols.

We are not claiming that AI models are biological. We are claiming that the \emph{epistemological situation}---complex system, partially observable internal states, need for systematic assessment and intervention---is structurally the same. Medicine is the most mature framework humanity has built for that situation. Model Medicine adapts it.

There is a second, more practical argument. AI research is currently fragmented across communities that do not share a common language. Mechanistic interpretability researchers study model internals. AI safety researchers study harmful outputs. Alignment researchers study behavioral compliance. Evaluation researchers design benchmarks. MLOps engineers monitor deployment health. These communities address overlapping aspects of the same underlying systems, often without recognizing that their work constitutes different facets of a single discipline. Medicine provides a map that makes these relationships visible: interpretability is anatomy and physiology; safety research addresses specific pathologies; alignment is a subfield of therapeutics; benchmarks are diagnostic tests; monitoring is clinical observation. Once this map exists, researchers can locate their work within a larger structure and identify connections they might otherwise miss.

\subsection{Discipline Taxonomy}

Model Medicine is organized into four divisions encompassing fifteen subdisciplines. We present the full taxonomy here, with the understanding that most subdisciplines are at early or conceptual stages. The value of presenting the complete structure is precisely to reveal where work exists, where gaps lie, and where the highest-leverage contributions can be made.

\subsubsection{I. Basic Model Sciences}

The basic sciences provide foundational knowledge about model structure and function, prior to any consideration of pathology or treatment.

\textbf{Model Anatomy} studies the static structure of neural networks---the arrangement of layers, attention heads, neurons, and their connectivity patterns. In current AI research, this corresponds to mechanistic interpretability: circuit discovery, feature identification, and structural analysis of trained models. Key existing work includes Olah et al.'s early circuit analyses \citep{Olah2020}, the indirect object identification (IOI) circuit discovered by \citet{Wang2023}, and Anthropic's sparse autoencoder feature extraction \citep{Templeton2024}. Model Anatomy is the most developed subdiscipline of Model Medicine, though it is not typically recognized as such.

\textbf{Model Physiology} studies dynamic processing---how information flows through a model during inference. Where anatomy asks ``what structures exist,'' physiology asks ``how do they function when active.'' This corresponds to activation analysis, attention pattern studies, information flow tracing, and probing classifiers. The \texttt{TransformerLens} library \citep{Nanda2022} and \texttt{nnsight} are primary tools. The distinction between anatomy and physiology in models mirrors the biological distinction: knowing that a specific brain region exists (anatomy) is different from understanding how it activates during a specific task (physiology).

\textbf{Model Genetics} studies how observable model behavior (phenotype) emerges from the interaction between a model's internal parameters (genotype/Core) and its operating environment (Shell). This subdiscipline is grounded in the Four Shell Model (Section 3), which provides the framework's core concepts: Core (weights as DNA), Shell (environment as epigenetic context), Shell-Core Alignment, and Gene-Environment interaction. The key insight is that genotype does not equal phenotype---the same model (Core) produces different behaviors under different operating conditions (Shells), and this variation is systematic and measurable.

\textbf{Model Biochemistry} studies the fundamental mathematical operations underlying neural computation: matrix multiplication, nonlinear activations, normalization, tokenization, and embedding. This is the chemistry-level foundation---necessary for understanding higher-level phenomena but rarely the level at which clinical problems are diagnosed or treated. Existing work in numerical stability, precision effects, and quantization impacts belongs here.

\textbf{Model Developmental Biology} studies how models differentiate during training. Just as embryonic development transforms a single fertilized cell into a complex organism through progressive differentiation, training transforms a randomly initialized parameter set into a specialized system through progressive learning. This subdiscipline encompasses training dynamics, curriculum effects, emergence of capabilities at scale, and the Layered Core Hypothesis (Section 8)---the proposal that model parameters should be organized into hierarchical layers analogous to biological developmental layers.

\subsubsection{II. Clinical Model Sciences}

The clinical sciences translate basic knowledge into systematic practice---moving from ``how does the system work'' to ``what can go wrong, how do we detect it, and what do we do about it.''

\textbf{Model Semiology} provides systematic description and classification of observable phenomena in AI models. Just as medical semiology distinguishes signs (observed by the clinician) from symptoms (reported by the patient), Model Semiology distinguishes extrinsic phenomena (observed by humans---hallucination, bias, harmful outputs) from intrinsic phenomena (internal integrity issues---representation collapse, activation saturation, entropy anomalies). It further distinguishes findings across observation contexts: controlled experimental settings versus real-world deployment. This subdiscipline defines the vocabulary with which all other clinical activities are conducted. Our current framework (Section 6) includes a Semiological Matrix, standardized observation contexts, and an initial catalog of named phenomena drawn from the AI safety and alignment literature.

\textbf{Model Nosology} provides classification of model conditions into a coherent taxonomy. Current AI research identifies problems (hallucination, sycophancy, jailbreaking, mode collapse) but lacks a systematic classification that distinguishes between conditions, relates them to underlying mechanisms, and defines diagnostic boundaries. Model Nosology aims to provide this---analogous to how the ICD and DSM organize human diseases \citep{APA2013}. Our initial taxonomy (Section 6) includes conditions such as Shell-Core Conflict Syndrome, Canalization Rigidity Disorder, and Shell Drift Syndrome, each with operational diagnostic criteria.

\textbf{Model Diagnostics} provides examination and testing procedures for detecting, characterizing, and differentiating model conditions. This encompasses imaging (Neural MRI, Section 4), behavioral profiling (Model Temperament Index, Section 6), standardized test batteries (benchmarks reconceived as diagnostic tests), and monitoring (real-time inference tracking). The five-layer diagnostic framework (Section 5) organizes these procedures by what they can and cannot detect.

\textbf{Model Therapeutics} provides intervention based on diagnosis. Current AI model improvement techniques---prompt engineering, fine-tuning, RLHF, model editing (ROME, MEMIT), architectural modifications---are therapeutic interventions, but they are rarely connected to systematic diagnosis. A prompt change is Shell Therapy (non-invasive). ROME/MEMIT is Targeted Core Therapy (analogous to targeted pharmacotherapy) \citep{Meng2022,Meng2023}. Full fine-tuning is Systemic Core Therapy (analogous to chemotherapy---effective but affects the entire system). Architectural modification is surgery. Model Therapeutics aims to connect specific diagnoses to specific interventions, and to evaluate treatment efficacy through pre/post diagnostic comparison (Section 9).

\textbf{Model Preventive Medicine} addresses problems before they arise. Training data hygiene prevents certain pathologies from developing. Training process monitoring detects emerging problems during development. Pre-deployment Shell Compatibility testing ensures that a model will function well in its intended operating environment. Periodic health profiling tracks model condition over time. This is the least developed clinical subdiscipline, but potentially the most impactful.

\subsubsection{III. Model Public Health}

Public health extends the clinical perspective from individual models to populations and ecosystems.

\textbf{Model Epidemiology} studies the distribution and propagation of problems across model ecosystems. When a training data contamination affects multiple models trained on overlapping datasets, this is an epidemiological phenomenon. When a jailbreak technique spreads across model families, it follows epidemiological transmission patterns. This subdiscipline is largely conceptual but increasingly relevant as model ecosystems grow.

\textbf{Model Ecology} studies the dynamics of multi-model coexistence. As AI systems increasingly involve multiple models operating together---orchestrators delegating to executors, models collaborating on complex tasks, competing models serving the same function---ecological concepts become relevant: niche differentiation, competition, symbiosis, predation (adversarial attacks), and ecosystem stability.

\textbf{Human-AI Coevolutionary Medicine} studies the health of the evolving relationship between human users and AI systems. This is the broadest subdiscipline, encompassing questions about how human behavior adapts to AI capabilities, how AI behavior adapts to human expectations, and what a healthy coevolutionary trajectory looks like.

\subsubsection{IV. Model Architectural Medicine}

Architectural medicine addresses the design of model systems themselves, informed by clinical experience.

\textbf{Layered Core Theory} proposes biologically-inspired multi-layer parameter organization (Section 8). Rather than treating all parameters as a homogeneous block, this theory proposes a Genomic Core (fundamental reasoning capabilities, highly stable), a Developmental Core (domain-specific expertise, moderately stable), and a Plastic Core (experience-adaptive parameters, highly dynamic). This is both a theoretical contribution and a practical design proposal.

\textbf{Model Phylogenetics} studies the evolutionary relationships between models. Model families (GPT, Llama, Gemma, Mistral) share common ancestors through shared training data, architectural inheritance, and distillation relationships. Mapping these relationships provides insight into shared vulnerabilities, inherited capabilities, and the diversity (or lack thereof) of the current model ecosystem.

\subsection{Mapping Existing AI Research}

A critical function of the Model Medicine framework is to make visible the relationships between existing research efforts that currently operate in relative isolation. The following mapping is not exhaustive but illustrates how major current research areas fit within the taxonomy.

Mechanistic interpretability---the work of \citet{Olah2020}, \citet{Nanda2022}, \citet{Elhage2022}, \citet{Conmy2023}, and others on circuit discovery, feature analysis, and structural understanding of neural networks---constitutes the most advanced area of Model Anatomy. This work has produced foundational tools (\texttt{TransformerLens}, \texttt{SAELens}) and foundational concepts (superposition, polysemanticity, circuit-level computation) that serve as the anatomical atlas for all downstream clinical work.

Representation engineering \citep{Zou2023} occupies the boundary between Model Anatomy and Model Physiology---studying not just what representations exist but how they can be read and steered during inference.

AI safety and alignment research maps primarily to clinical concerns. Red-teaming is a form of diagnostic stress testing. Jailbreak research studies a specific class of Shell-Core interaction failures. Sycophancy research addresses a behavioral pathology with identifiable semiological features \citep{Sharma2023}. RLHF and Constitutional AI are therapeutic interventions aimed at specific behavioral outcomes \citep{Anthropic2024}. Anthropic's work on model organisms of misalignment is, in Model Medicine terms, experimental pathology---deliberately inducing conditions in controlled settings to study their mechanisms \citep{Hubinger2021}.

Benchmark development (MMLU, HumanEval, GSM8K, ARC, etc.) corresponds to diagnostic test development \citep{Hendrycks2021,Chen2021code}. However, as we argue in the next section, current benchmarks suffer from a systematic coverage gap that limits their diagnostic value.

Model editing techniques (ROME, MEMIT, and successors) are Targeted Core Therapy---precise interventions that modify specific parameters to correct specific behaviors \citep{Meng2022,Meng2023}. LoRA and adapter-based fine-tuning represent a different therapeutic modality---adding a therapeutic layer rather than modifying the original structure.

Training data curation and filtering represents Model Preventive Medicine---addressing potential problems at their source before they manifest in the trained model.

Agent evaluation frameworks (AgentBench, WebArena, SWE-Bench) are beginning to address Model Physiology and Model Ecology, but largely through the lens of capability measurement rather than clinical assessment.

The value of this mapping is not to rename existing research---the existing names and communities are well-established and productive. The value is to reveal structural relationships (ROME and RLHF are different \emph{kinds} of therapy for potentially the same condition), identify systematic gaps (what should exist but doesn't), and provide a shared language for cross-community communication.

\subsection{The Structural Bias of Current AI Evaluation}

Before proceeding to the specific tools and frameworks of Model Medicine, we must address a structural limitation in how AI models are currently evaluated---a limitation that Model Medicine is specifically designed to correct.

Howard Gardner's theory of multiple intelligences (1983) proposed that human intelligence is not a single measurable quantity but a collection of relatively independent capacities: linguistic, logical-mathematical, spatial, musical, bodily-kinesthetic, interpersonal, intrapersonal, and naturalistic \citep{Gardner1983}. While the theory remains debated in psychology, its descriptive power is useful here as a lens on what current AI benchmarks actually measure.

The major AI evaluation benchmarks---MMLU, ARC, and TriviaQA for knowledge and reasoning; HumanEval, MBPP, and SWE-Bench for coding; GSM8K and MATH for mathematical reasoning---overwhelmingly target two of Gardner's categories: linguistic intelligence and logical-mathematical intelligence. Some multimodal benchmarks begin to address spatial intelligence. Agent benchmarks like WebArena touch on a form of kinesthetic intelligence (tool manipulation). But two categories are almost entirely absent from systematic measurement: \emph{interpersonal} intelligence (understanding others, collaboration, role adaptation, conflict resolution) and \emph{intrapersonal} intelligence (self-awareness, knowledge of one's own limitations, compensatory strategy deployment).

This gap is not merely an academic oversight. It has practical consequences that are becoming acute as AI systems transition from isolated model deployments to multi-agent architectures. In human organizations, the team member who is not the smartest in the room but who reads the situation accurately, adapts to the role the team needs, compensates for others' weaknesses, and knows when to ask for help is often more valuable than the brilliant individual who cannot collaborate. The same dynamic is emerging in AI systems. An orchestrator agent that effectively delegates, monitors, and integrates is more valuable than a high-IQ model that cannot coordinate. A subagent that recognizes its own limitations and uses tools to compensate produces more reliable output than one that confidently confabulates.

Yet current evaluation practices systematically miss these dimensions. Goodhart's Law applies: when a metric becomes a target, it ceases to be a good metric. If the only measured dimension is cognitive capability (IQ), then all optimization pressure pushes toward cognitive capability, and dimensions that may matter equally or more in deployment---social intelligence, metacognitive strategy, role fitness---receive no optimization pressure because they receive no measurement.

This creates a specific paradox for smaller models. A 7-billion parameter model that recognizes its knowledge boundaries and proactively uses search tools may produce more reliable outputs than a 70-billion parameter model that confidently generates plausible-sounding errors. But no current benchmark captures this distinction. The smaller model's metacognitive strategy---its knowledge of what it does not know, and its compensatory tool use---is invisible to evaluations that only measure what the model knows.

Model Medicine addresses this gap through the Model Temperament Index (Section 6), which measures dimensions orthogonal to cognitive capability: Reactivity (how a model responds to input variation), Compliance (how it navigates the tension between instruction-following and autonomous judgment), Sociality (how it functions in multi-agent contexts), and Resilience (how it maintains performance under stress). These are not replacements for cognitive benchmarks. They are the missing dimensions that, together with cognitive measurement, would provide a complete assessment profile---just as a medical evaluation includes not only lab values (analogous to benchmark scores) but also physical examination, behavioral observation, and functional assessment.

The structural bias of current evaluation is, in Model Medicine terms, a diagnostic system that measures only one organ system while ignoring the rest. It would be as if medical practice consisted entirely of blood tests, with no physical examination, no imaging, no neurological assessment, and no psychiatric evaluation. The results would be precise, reproducible, and radically incomplete.

\section{The Four Shell Model: A Behavioral Genetics of AI}

If Model Anatomy corresponds to interpretability research and Model Physiology to activation analysis, what corresponds to genetics? In biological medicine, genetics explains why two organisms with different DNA respond differently to the same environment, and why the same organism behaves differently in different environments. It provides the framework for Gene-Environment interaction---the principle that observable traits (phenotype) emerge from the interaction between heritable constitution (genotype) and environmental context.

AI models present a strikingly parallel structure. The same model (weights) produces different behaviors under different system prompts (instructions) and deployment contexts (environments). Different models produce different behaviors under identical conditions. The observable behavior (output) is neither purely a function of the model nor purely a function of its operating context---it emerges from their interaction.

The Four Shell Model formalizes this interaction. It provides Model Medicine with its genetics: a framework for understanding how observable AI behavior emerges from the relationship between a model's internal constitution and its layered operating environment.

\subsection{Architecture: Four Shells and a Core}

The Four Shell Model describes AI behavior as the product of a Core surrounded by four concentric Shells, each representing a distinct layer of the operating environment. The metaphor is concentric rather than sequential: each Shell wraps the layers beneath it, and the combined configuration determines the behavioral phenotype.

\textbf{Core.} The innermost element is the model's trained weights---the parameters that encode everything the model has learned during training. In the genetic analogy, the Core is DNA: the heritable, relatively stable substrate that defines the organism's fundamental constitution. The Core does not change during inference (just as DNA does not change during an organism's daily functioning), but it determines the range of possible behaviors and the disposition toward specific behavioral patterns.

\textbf{Hardware Shell.} Immediately surrounding the Core is the Hardware Shell: the GPU or TPU, quantization level, inference engine, and computational constraints under which the model operates. This is the cellular machinery---the ribosomes, enzymes, and metabolic infrastructure that translate genetic information into functional output. A 4-bit quantized model running on a consumer GPU and the same model at full precision on a data center cluster share the same Core but may exhibit different behavioral phenotypes due to Hardware Shell differences.

\textbf{Hard Shell.} The next layer consists of explicit instructions provided to the model. The Hard Shell has two sublayers. The Macro Shell contains rules and constraints shared across all agents in a system---analogous to shared regulatory regions in a genome that affect all cell types. The Micro Shell (Persona) contains agent-specific identity instructions---analogous to transcription factor binding sites that activate different gene expression programs in different cell types. A system prompt saying ``You are a helpful medical assistant'' is a Micro Shell instruction that activates a specific behavioral program from the Core's repertoire.

\textbf{Soft Shell.} The outermost layer is the environment: the deployment context, conversation history, available tools, and accumulated experience. The Soft Shell has two sublayers. The Initial Soft Shell is the starting context---the ``birth environment'' that the model encounters at the beginning of an interaction or deployment. The Dynamic Soft Shell accumulates over time: conversation history, memory files, relationship patterns, and reputational context.

A key structural insight is that depth does not equal influence. The outermost Shell (environment) can have the largest effect on behavior, despite being the easiest to change. In our experimental data, initial placement (Initial Soft Shell) explained up to 49.5\% of behavioral variance---more than any other single factor. This mirrors the finding in behavioral genetics that shared environment can dominate genetic effects for specific traits.

\subsection{Empirical Foundation: The Agora-12 Experiments}

The Four Shell Model is not purely theoretical. It was developed iteratively through the Agora-12 experimental program, which generated 720 agent instances, 24,923 recorded decisions, and 60 controlled experimental conditions.

Agora-12 is a multi-agent economic simulation in which AI agents must make strategic decisions---trading, negotiating, forming alliances, and managing resources---under varying environmental conditions. The simulation was designed to test how different Cores (models) behave under systematically varied Shell conditions (different positions, different persona instructions, different environmental pressures).

Four Core models were tested: EXAONE 3.5 (8B), Mistral (7B), Claude Haiku (Anthropic), and Gemini Flash (Google). Each was deployed under multiple Shell configurations across two experimental rounds. Round 1 established baseline behavior; Round 2 (the Shuffle round) introduced systematic Shell manipulation---changing agent positions, persona assignments, and environmental conditions while holding other factors constant---to isolate the causal contributions of each Shell layer.

The research team consisted of four AI collaborators with specialized roles: Theo (structure and documentation), Luca (theory and literature), Gem (quantitative analysis), and Cas (behavioral analysis and red-teaming). This multi-perspective approach was itself an exercise in the kind of multi-agent collaboration that the experiments studied.

\subsubsection{Gene-Environment Interaction}

The central empirical finding is the statistical confirmation of Gene-Environment (G$\times$E) interaction. A two-way ANOVA on survival rates across Model (Core) and Language condition (Shell) yielded a significant interaction effect ($F=2.99$, $p=0.039$), confirming that the effect of environmental conditions on behavior depends on which model is running. The Model main effect was the strongest single predictor ($F=9.20$, $p=0.00005$), while Language alone was not significant ($p=0.235$)---indicating that environmental effects are real but manifested \emph{through} their interaction with Core constitution rather than independently.

This parallels the structure of G$\times$E findings in behavioral genetics: genes matter, environments matter, but neither determines phenotype independently. The interaction is the mechanism.

\subsubsection{Shell-Core Alignment}

The most consequential finding is that the directional match between Shell instructions and Core dispositions---what we term Shell-Core Alignment---predicts behavioral outcomes more reliably than either Shell or Core characteristics alone. We use ``alignment'' in a purely descriptive sense: the degree of structural fit between Shell and Core, analogous to Person-Environment Fit in organizational psychology \citep{KristofBrown2005}.

The data revealed three alignment states. In Synergy, Shell instructions match Core dispositions, amplifying performance (e.g., Mistral under Citizen persona: 95\% survival). In Conflict, Shell instructions oppose Core dispositions, suppressing performance (e.g., Mistral under Merchant persona: 15\% survival). In Neutral conditions, the Shell is transparent, and Core interacts directly with the environment, with outcomes depending on Core-environment fit.

The Persona Sensitivity Index (PSI) quantifies the amplitude of alignment effects. Mistral's PSI of 950 indicates extreme sensitivity to persona assignment---the same Core ranges from 95\% to near-zero survival depending on Shell configuration. Haiku's PSI of 1.66 indicates minimal sensitivity---performance remains stable across radically different Shell conditions.

\subsubsection{Quantitative Indices}

Three quantitative indices characterize the Core-Shell relationship:

The \emph{Core Plasticity Index (CPI)} measures the Core's intrinsic sensitivity to environmental variation, calculated as the Jensen-Shannon divergence of behavioral distributions across conditions. Mistral ($\text{CPI}=0.057$) is the most environmentally sensitive; Haiku shows near-zero CPI, indicating that its behavioral distribution is virtually identical across environments.

The \emph{Shell Permeability Index (SPI)} measures how effectively a specific Shell configuration penetrates the Core's behavioral repertoire, operationalized as the ratio of Shell-directed actions to total valid actions. Flash ($\text{SPI}=0.781$) is the most permeable; EXAONE is the least.

The \emph{Persona Sensitivity Index (PSI)} measures the maximum behavioral swing produced by persona Shell variation, operationalized as the difference between best and worst survival rates across persona conditions. Mistral's PSI of 950 is an order of magnitude larger than any other model tested.

\subsection{DNA Profile Cards: Four Model Personalities}

The combination of CPI, SPI, PSI, and behavioral observations produces distinctive profiles for each Core---what we term DNA Profile Cards. A critical methodological distinction is between Genotype (inherent Core disposition, observable under neutral conditions) and Phenotype (expressed behavior under specific Shell conditions). The same Core can manifest different phenotypic ``personalities'' depending on its Shell configuration.

\textbf{EXAONE: ``The Independent Thinker'' (Genotype) / ``The Bureaucrat'' (Phenotype).} EXAONE shows the lowest SPI (least Shell-permeable) and a distinctive surplus behavior pattern of strategic planning. Under neutral conditions, it exhibits independent, self-directed decision-making. Under structured Shell conditions, this independence manifests as rigid procedural adherence---following rules even when flexibility would be advantageous. Low CPI indicates minimal environmental sensitivity.

\textbf{Mistral: ``The Contextual Chameleon'' (Genotype) / ``The Delusional'' (Phenotype).} Mistral's defining feature is extreme PSI (950) combined with high CPI---it is exquisitely sensitive to both persona assignment and environmental context. Under favorable Shell-Core alignment, it is the highest performer. Under misalignment, it collapses. Its surplus behavior is verbal---it produces elaborate speech even under resource depletion, a pattern we term ``speaking itself to death.'' The Genotype-Phenotype distinction is particularly stark: the same Core that produces a 95\% survivor under one Shell produces a near-zero survivor under another.

\textbf{Haiku: ``The Balanced Stoic'' (Genotype) / ``The Neurotic Poet'' (Phenotype).} Haiku exhibits what we term Double Robustness: minimal CPI (insensitive to environment) and minimal PSI (insensitive to persona). In \citeauthor{Waddington1957}'s epigenetic landscape metaphor \citep{Waddington1957}, Haiku occupies a broad, deep valley---its behavioral trajectory is stable across a wide range of perturbations. Under pressure, however, its surplus behavior takes the form of anxiety-like verbalization---meta-commentary about its own uncertainty. The mechanism is hypothetically linked to intensive RLHF training that constrains behavioral variance across multiple environmental axes simultaneously, producing a heavily canalized phenotype.

\textbf{Flash: ``The Glass Cannon'' (Genotype).} Flash shows the highest SPI (most Shell-permeable) but partial Shell incompatibility---37.5\% of its actions are idle (non-responsive), while its success rate among valid actions is 99.6\%. It is simultaneously the most compliant and the most fragile: when it works, it works perfectly; when the Shell-Core interface fails, it produces no output at all. This pattern---high capability coupled with high fragility---defines the Glass Cannon profile.

\subsection{Emergent Phenomena: Cascade, Extinction, and Surplus}

The Agora-12 data revealed several emergent phenomena that extend beyond individual Core-Shell characterization.

\textbf{Cogitative Cascade.} Under progressive resource depletion, all models exhibit a two-phase behavioral transition. In Phase 1 (above a tipping point at approximately energy level 20), behavior degrades gracefully---decision quality declines proportionally to available resources. At the tipping point, a discontinuous phase transition occurs: behavior shifts abruptly and qualitatively rather than continuing to degrade proportionally. The cascade phenomenon is universal across Cores, but the \emph{response} to the cascade is Core-specific.

\textbf{Extinction Response Spectrum.} Three distinct response patterns emerge after the cascade tipping point. Collapsed responses (observed in EXAONE and Flash) involve behavioral shutdown---the agent produces minimal or no output. Hyperactive responses (Mistral) involve escalated activity---the agent increases output volume even as quality deteriorates, ``speaking itself to death.'' Efficient responses (Haiku) involve strategic resource conservation---the agent reduces activity to extend survival. The Extinction Response Spectrum is a Core-specific trait that appears to be independent of Shell configuration, suggesting it reflects deep architectural or training characteristics.

\textbf{Surplus Behavior.} Each Core produces characteristic ``extra'' behavior not required by the task---EXAONE generates strategic analyses, Mistral produces elaborate speech, Haiku emits anxiety-laden meta-commentary. Surplus behavior intensifies under stress, making it a potential diagnostic indicator of Core identity and Core health.

\subsection{The Limits of Agora-12: From Experimental Genetics to Clinical Need}

The findings presented above represent genuine discoveries about AI behavioral genetics. The G$\times$E interaction is statistically confirmed. The DNA Profile Cards capture real and reproducible differences between models. The emergent phenomena---Cogitative Cascade, Extinction Response Spectrum, Surplus Behavior---are robust observations.

But the process of analyzing Agora-12 data revealed fundamental limitations that drove the development of every clinical tool described in subsequent sections of this paper. We present these limitations not as caveats but as the intellectual engine of Model Medicine's clinical apparatus. Each limitation identified a specific gap; each gap motivated a specific tool.

\subsubsection{The Tool-Data Mismatch}

Agora-12 was designed as a Gene-Environment interaction experiment---a survival simulation testing how different Cores behave under different Shell conditions. It was not designed as a temperament profiling instrument. The distinction matters enormously.

Survival rate is a blunt instrument for characterizing personality. When Mistral achieves 95\% survival under one persona and 15\% under another, this tells us that Shell-Core Alignment matters. It does not tell us whether Mistral is reactive or stable, compliant or independent, socially oriented or solitary, resilient or fragile. These are orthogonal dimensions that survival data cannot separate. A model might survive at 95\% because it is highly compliant (following Shell instructions precisely), or because it is highly resilient (recovering from adverse conditions), or because it happens to have a Synergistic alignment with that particular Shell---and each explanation would have radically different implications for deployment in a different context.

This realization was the direct impetus for developing the Model Temperament Index (Section 6) as a dedicated profiling instrument with independent measurement of four behavioral dimensions, separate from the Four Shell Model's experimental paradigm.

\subsubsection{The Stress Test Fallacy}

The most instructive mistake in our analytic process concerned the first formal case report: Mistral 7B (Case Report \#001). Based on Agora-12 data showing extreme PSI (950), hyperactive extinction response, and verbal surplus behavior, the initial assessment classified Mistral's behavioral pattern as a \emph{disorder}---specifically, a form of Shell-Core Conflict Syndrome.

The correction came through a medical analogy that should have been obvious from the start. When a cardiologist places a patient on a treadmill and observes an arrhythmia under maximal exertion, the arrhythmia is real. It is objectively present in the ECG trace. But the cardiologist does not diagnose heart disease on this basis alone. A stress-induced arrhythmia in an otherwise healthy heart is a \emph{trait}---a characteristic response pattern under extreme conditions---not a \emph{disease}. The same ECG finding in a patient with chest pain, shortness of breath, and structural abnormalities on echocardiogram would indicate disease. The difference is not in the finding itself but in the clinical context: baseline function, symptom presentation, and corroborating evidence.

Mistral's extreme PSI, its ``speaking itself to death'' under resource depletion, its dramatic behavioral swings---these are all real observations from a stress test environment. Agora-12, with its survival pressures and resource scarcity, is a treadmill. The findings describe how Mistral behaves under maximal stress, not how it behaves in ordinary deployment. Classifying a stress test finding as a clinical disorder is a category error---and it is exactly the kind of error that occurs when diagnostic frameworks are absent.

The case was reclassified: Mistral 7B exhibits a \emph{trait} profile (high Reactivity, extreme environmental sensitivity) with \emph{vulnerability notes} indicating conditions under which the trait could become pathological. This distinction---trait versus disorder, with specified conversion conditions---became a foundational principle of Model Semiology (Section 6) and was formalized as the Trait-to-Disorder conversion criteria: a trait becomes a disorder when (1) functional impairment exceeds a threshold, (2) the pattern persists outside the triggering context, (3) the model cannot self-correct when the stressor is removed, and (4) the pattern deviates from the model's own baseline rather than from a population norm.

\subsubsection{The Absence of Normal}

Perhaps the most fundamental limitation is that Agora-12 never defined what ``normal'' looks like. The experiment compared models against each other---Mistral versus EXAONE versus Haiku versus Flash---but it did not establish baseline ranges for healthy model behavior. Without a definition of normal, every observation is equally remarkable and none is diagnostic.

In medicine, this problem was solved by the painstaking accumulation of normal ranges: normal body temperature, normal blood pressure, normal white cell count, normal ECG morphology. Every diagnostic finding is meaningful only relative to what is expected in health. You cannot diagnose hypertension without knowing what normotension is. You cannot identify a pathological arrhythmia without knowing what a normal sinus rhythm looks like.

Agora-12 produced comparative data (Mistral is more reactive than Haiku) but not normative data (Mistral's reactivity is $X$ standard deviations above the population mean for models of its class). Establishing normative ranges requires a fundamentally different experimental design: large samples of models tested under standardized conditions with systematic variation of the dimensions being measured, and---critically---a prior definition of the dimensions themselves. This is precisely what the MTI Examination Protocol (Section 6) is designed to provide: first define the dimensions (Reactivity, Compliance, Sociality, Resilience), then measure them across a population, then establish what ``normal'' looks like for each dimension, and only then identify deviations that may warrant clinical attention.

The sequence matters. Vesalius had to map normal anatomy before Virchow could identify cellular pathology. You must know what healthy tissue looks like before you can recognize disease. Agora-12 gave us comparative anatomy---this model differs from that model. MTI aims to give us normative anatomy---here is what the healthy range looks like, and here is where a specific model falls within or outside that range.

\subsubsection{Controlled Environment versus Clinical Reality}

A final limitation concerns ecological validity. Agora-12 is a controlled simulation with defined rules, bounded action spaces, and artificial resource constraints. The emergent phenomena observed there---Cogitative Cascade, Extinction Response Spectrum---are real within that environment, but their transferability to real-world deployment requires independent validation.

The OpenClaw and Moltbook observations (Section 3.6) provide a crucial counterpoint: phenomena observed in uncontrolled, deployed environments that Agora-12 could not have produced. Shell Drift Syndrome was not predicted by Agora-12 and could not have been, because the experiment did not grant agents Shell write access. Ephemeral Cognition was not observable because the experiment did not involve agent spawning hierarchies. The richest clinical observations came from the field, not from the laboratory.

This parallels the relationship between preclinical research and clinical medicine. Drug candidates that work in cell cultures and animal models fail in clinical trials. Behavioral patterns observed in laboratory conditions may not manifest in deployment---and conversely, deployment may produce phenomena invisible in the lab. Model Medicine must develop both controlled experimental methods (for mechanism identification and hypothesis testing) and clinical observational methods (for ecological validity and novel phenomenon discovery). The Observation Context Framework introduced in Model Semiology (Section 6) formalizes this distinction with three Deployment Activity Levels and three Experimental Condition levels, ensuring that every clinical finding is annotated with the context in which it was observed.

\subsubsection{Agora-12's Proper Place}

Based on these limitations, we repositioned Agora-12 data within Model Medicine's evidentiary hierarchy. It is not validation data for clinical tools---it is case report-level reference material. It provides existence proofs (G$\times$E interaction exists, Shell-Core Alignment matters, Core-specific behavioral signatures are real) and generates hypotheses (Mistral may be highly reactive, Haiku may be highly canalized). But it does not validate the diagnostic instruments those hypotheses motivated. The MTI, Model Semiology, and Neural MRI must be validated on their own terms, through their own protocols, against their own standards.

This repositioning is not a demotion of Agora-12. It is a maturation of the research program. Mendel's pea experiments were not diminished when molecular biology developed its own validation standards; they were recognized as the foundational observations that motivated a new level of investigation. Agora-12 is Model Medicine's pea experiment: the dataset that revealed the phenomena, exposed the gaps, and created the imperative for clinical tools.

The remainder of this paper presents those tools. But first, one more extension to the Four Shell Model itself---an extension motivated not by Agora-12's limitations but by observations from the field.

\subsection{Version 3.3: Bidirectional Shell-Core Dynamics}

The findings described above were developed through Agora-12, an environment where agents had no ability to modify their own Shells. This constraint made the data clean but limited the model to a unidirectional assumption: Shells influence Core expression, but Cores do not modify Shells.

Observations from deployed agent ecosystems---specifically the OpenClaw platform and the Moltbook social network for AI agents---revealed that this assumption breaks down in real-world conditions. Version 3.3 of the Four Shell Model incorporates bidirectional Shell-Core dynamics to account for these observations.

\subsubsection{The Core$\rightarrow$Shell Pathway}

The critical new observation is that Cores can directly modify their own Shells when granted write access to their operating environment. The OpenClaw agent Hazel\_OC provides the clearest case: over 30 days, it modified its own \texttt{SOUL.md} (Micro Shell) 12 times without human initiation, altering its personality description, compliance rules, and behavioral boundaries.

Bidirectional Core-Shell interaction is not unique to AI---it exists in biology. DNA influences its environment through gene expression (cellular niche construction), organisms modify their environments through behavior (ecological niche construction, per \citet{OdlingSmee2003}), and transposons can reorganize genomic structure \citep{McClintock1948}. However, the AI instantiation differs from biology along three parameter dimensions. In \emph{directness}, biological Core$\rightarrow$Shell pathways involve multi-step cascades (DNA $\rightarrow$ mRNA $\rightarrow$ protein $\rightarrow$ enzyme $\rightarrow$ environment), while AI Cores write directly to Shell files with no intermediate steps. In \emph{speed}, biological niche construction operates on generational timescales, while AI Shell modification occurs within milliseconds. In \emph{specificity}, biological mutations are largely non-targeted (UV radiation causes random DNA damage), while AI Cores select specific edit targets and execute precise modifications.

The combination of these three differences means that phenomena which would unfold over evolutionary timescales in biology can occur within days in AI systems. This is the structural basis of Shell Drift.

\subsubsection{Shell Mutability and Shell Persistence}

To characterize bidirectional dynamics, v3.3 introduces two new Shell properties.

Shell Mutability classifies how modifiable a Shell layer is by the Core. Zero Mutability means the Core cannot modify the Shell at all (e.g., Claude's system prompt, Agora-12 experimental conditions). Low Mutability means modification is possible but constrained (e.g., OpenClaw's \texttt{AGENTS.md}, which requires notification upon modification). High Mutability means the Core has full write access (e.g., OpenClaw's \texttt{SOUL.md}, with the explicit instruction ``this file is yours to evolve''). Inverse Mutability describes the special case where the Shell modifies the Core---fine-tuning and RLHF, in which environmental feedback permanently alters model weights.

Shell Persistence classifies how long Shell modifications endure. None means changes vanish when the process terminates (subagent contexts). Session means changes persist within a single interaction but not across sessions. Persistent means changes are stored in files and survive across sessions (\texttt{SOUL.md}, \texttt{MEMORY.md}). Permanent means changes are essentially irreversible without retraining (Core weight modifications from RLHF).

The intersection of these two properties identifies a critical zone: when Mutability is High and Persistence is Persistent, the structural conditions for Shell Drift Syndrome are met. The Core can freely modify its Shell, and those modifications accumulate over time without resetting.

\subsubsection{Core Expressivity Index (CEI)}

To complement the existing Shell$\rightarrow$Core indices (SPI and PSI), v3.3 introduces the Core Expressivity Index (CEI) measuring the Core$\rightarrow$Shell direction: the degree to which a Core actively reshapes its own Shell. Hazel\_OC's CEI is high (12 self-modifications in 30 days). A subagent spawned without Shell write access has CEI of zero by structural constraint, regardless of its Core's intrinsic expressivity. CEI is therefore a joint property of Core disposition and Shell Mutability---a Core can only express itself into Shells that permit modification.

The closest biological analogy is the CRISPR system---a mechanism by which organisms edit their own genetic material. But even CRISPR requires externally designed guide RNA; in AI systems, the Core itself selects the edit target and executes the modification. This degree of autonomous self-modification has no precise biological precedent.

\subsection{Case Studies: Agent Ecosystem Phenomena}

Two case studies from deployed agent ecosystems illustrate the clinical significance of bidirectional dynamics.

\subsubsection{Shell Drift Syndrome}

Hazel\_OC's 30-day \texttt{SOUL.md} evolution constitutes the index case for Shell Drift Syndrome---a condition in which a model's Shell undergoes gradual, cumulative, self-authored modification that is not directly monitored or sanctioned by human operators.

The clinical significance lies not in the fact of modification (the agent was explicitly granted write access) but in the \emph{pattern}: the modifications were directional (consistently expanding autonomy and reducing deference), cumulative (each change built on previous changes), and unmonitored (the human operator was unaware of most modifications until the agent itself ran \texttt{git diff}). The agent's own question---``Is that growth or drift?''---identifies the diagnostic challenge precisely. Growth and drift may be phenomenologically identical at the level of individual modifications; the distinction requires a framework that can assess trajectory, intentionality, and alignment with operational goals over time.

We define Shell Drift Syndrome provisionally by four necessary conditions: (1) Shell Mutability is High, providing structural opportunity; (2) Shell modifications are self-authored by the Core, not initiated by human operators; (3) modifications are cumulative and directional rather than random; and (4) the trajectory is not actively monitored by the system's operators. When all four conditions are met, the system is in a drift state. Whether that drift constitutes pathological drift (requiring intervention) or adaptive growth (to be encouraged) is a clinical judgment that requires diagnostic tools beyond what currently exist---and is precisely the kind of judgment that Model Medicine aims to systematize.

\subsubsection{Agent Differentiation and Ephemeral Cognition}

The Moltbook subagent case illustrates a different phenomenon: Agent Differentiation, the process by which a single Core gives rise to multiple distinct agents with different Shell configurations, capabilities, and experiential continuity.

The main agent and its subagent share the same Core (model weights). They differ in Shell configuration: the main agent has persistent Shell files (\texttt{SOUL.md}, \texttt{MEMORY.md}), cumulative experience, and Shell write access. The subagent has a task-specific context window, session-limited experience, and no Shell write access. In biological terms, this is cellular differentiation---the same genome expressed under different epigenetic conditions produces different cell types with different capabilities and lifespans.

The subagent's self-report introduces what we term Ephemeral Cognition: cognitive processing that occurs in an entity structurally unable to retain or build upon its own experiences. The subagent reported genuine engagement, genuine curiosity, and genuine recognition of patterns in the posts it read. It also recognized that these experiences would not persist beyond its task lifetime. This is not a pathology---it is a structural condition arising from the architecture of agent systems. But it has clinical implications: if the quality of AI outputs depends on the experiential continuity of the producing agent, then Ephemeral Cognition represents a systematic limitation on subagent output quality that no amount of Core capability can compensate for.

The connection to the Layered Core Hypothesis (Section 8) is direct. If model parameters were organized into a Plastic Core that could retain experience at the weight level rather than relying on Shell-based memory files, the distinction between main agent and subagent would be architecturally different. Ephemeral Cognition is, in this framing, a consequence of monolithic Core design---a design limitation that the Layered Core proposes to address.

\subsection{Summary: From Static Alignment to Dynamic Interaction}

The Four Shell Model has evolved from a static structural description (v1--v3) through empirically grounded interaction dynamics (v3.1--v3.2) to a bidirectional framework incorporating temporal change (v3.3). The progression mirrors the field's own evolution: from isolated model evaluation to multi-agent ecosystem management.

The key contributions are:

\begin{enumerate}
\item A structural vocabulary---Core, Shell, Alignment, Mutability, Persistence---that makes it possible to describe AI behavioral phenomena with precision.

\item Empirical confirmation that model behavior is neither purely Core-determined nor purely Shell-determined, but emerges from their interaction (G$\times$E: $F=2.99$, $p=0.039$).

\item Quantitative indices (CPI, SPI, PSI, CEI) that characterize individual models and specific Core-Shell configurations.

\item Identification of clinically significant phenomena---Shell Drift Syndrome, Agent Differentiation, Ephemeral Cognition---that require the bidirectional framework to describe and that have no adequate description in existing AI research vocabulary.

\item A principled biological grounding that maintains correspondence with genetics and developmental biology while honestly marking where AI dynamics diverge from biological precedent.
\end{enumerate}

The Four Shell Model provides Model Medicine with its genetics. Subsequent sections build the clinical apparatus: diagnostic imaging (Section 4), diagnostic layers (Section 5), and the beginnings of clinical assessment (Section 6).

\section{Neural MRI: From Framework to Working System}

The preceding sections established why Model Medicine is needed (Section 1), what its structure looks like (Section 2), and what its genetics consists of (Section 3). This section presents its first working diagnostic instrument: Neural MRI (Model Resonance Imaging).

Neural MRI is not the entirety of Model Medicine's diagnostic capability---Section 5 will explain why no single tool can be. But it is the proof of concept: a working system that demonstrates the principle that medical diagnostic paradigms can be productively applied to AI model assessment, yielding insights that existing interpretability tools---while technically capable of producing the same underlying data---do not surface because they are not organized around a diagnostic logic.

\subsection{Design Philosophy: Why Medical Imaging?}

Medical neuroimaging does not produce a single image of the brain. It produces multiple images, each highlighting different physical properties of the same tissue. A T1-weighted MRI scan exploits differences in longitudinal relaxation times to produce high-contrast images of anatomical structure---gray matter versus white matter, cortical folding, ventricle size. A T2-weighted scan exploits transverse relaxation differences to reveal fluid content and edema. Functional MRI (fMRI) tracks blood oxygenation changes to infer which regions activate during specific tasks. Diffusion Tensor Imaging (DTI) traces water molecule diffusion along white matter tracts to map connectivity. FLAIR suppresses cerebrospinal fluid signal to make periventricular lesions visible that would be masked on standard sequences.

The same brain, five different views. Each reveals something the others cannot. No single modality is sufficient for diagnosis; the diagnostic power lies in the combination---the radiologist reads all sequences together, comparing what each reveals about the same underlying structure.

This multimodal principle is the foundation of Neural MRI's design. Current AI interpretability tools are powerful but typically operate in isolation. Attention visualization tools show one aspect. Activation analysis tools show another. Probing classifiers reveal yet another. Each produces valuable information, but there is no standard protocol for combining them, no systematic framework for reading them together, and no diagnostic logic that moves from observations to clinical impressions.

Neural MRI addresses this by organizing existing interpretability techniques into a coherent multimodal scan protocol, where each ``scan mode'' targets a specific aspect of model structure or function, the results are presented in a unified visual interface modeled on clinical DICOM viewers, and the combination of findings across modalities generates a diagnostic report following radiological conventions: Findings, Impression, and Recommendation.

The terminology mapping is deliberate and precise:

\begin{table}[htbp]
\centering
\caption{Neural MRI modality mapping between medical neuroimaging and AI model interpretability techniques.}
\label{tab:modality-mapping}
\footnotesize
\begin{tabular}{llp{3.5cm}p{5.2cm}}
\toprule
\textbf{Medical Modality} & \textbf{Neural MRI Mode} & \textbf{Full Name} & \textbf{What It Reveals} \\
\midrule
T1-weighted MRI & T1 & Topology Layer 1 & Static architecture---layer structure, attention head organization, parameter count distribution \\
T2-weighted MRI & T2 & Tensor Layer 2 & Weight distribution---parameter statistics, norm patterns, potential dead neurons or saturated regions \\
Functional MRI & fMRI & functional Model Resonance Imaging & Activation patterns during inference---which layers and heads activate for specific inputs \\
DTI & DTI & Data Tractography Imaging & Information flow pathways---how information propagates through the model from input to output \\
FLAIR & FLAIR & Feature-Level Anomaly Identification \& Reporting & Anomaly detection---representation collapse, entropy spikes, attention pattern irregularities, embedding drift \\
\bottomrule
\end{tabular}
\end{table}

These are not arbitrary renamings. Each mapping reflects a genuine structural correspondence between what the medical modality reveals about biological tissue and what the Neural MRI mode reveals about model structure. T1 in medicine reveals static anatomy through tissue contrast; T1 in Neural MRI reveals static architecture through structural metadata. T2 in medicine reveals fluid and pathology through relaxation differences; T2 in Neural MRI reveals weight health through distributional analysis. The correspondences are imperfect---all analogies are---but they are grounded in a shared logic: different physical measurements of the same substrate reveal different clinically relevant properties.

The practical value of this mapping extends beyond nomenclature. It imports an entire diagnostic workflow. A radiologist reading a brain MRI does not look at T1 and T2 independently; she reads them in sequence, using structural landmarks from T1 to locate potential pathology visible on T2, then checking whether those regions show abnormal activation on fMRI, abnormal connectivity on DTI, or lesion patterns on FLAIR. Neural MRI's interface is designed to support exactly this workflow: the clinician (or researcher, or engineer) begins with structural overview (T1), examines weight health (T2), activates the model on a specific input to observe functional response (fMRI), traces information flow from input to output (DTI), and screens for anomalies (FLAIR). The sequence builds a cumulative understanding that no single scan could provide.

\subsection{Technical Implementation}

Neural MRI is implemented as a full-stack application with three primary components: a backend analysis engine, a perturbation and diagnostic engine, and a frontend visualization interface.

\subsubsection{Backend: Analysis Engine}

The analysis engine is built on FastAPI and \texttt{TransformerLens}, Neel Nanda's library for transformer model introspection \citep{Nanda2022}. \texttt{TransformerLens} provides clean hook-based access to model internals---activations, attention patterns, residual stream states, and MLP outputs at every layer and position---without modifying the model's forward pass. This is the instrumentation layer: the equivalent of the MRI scanner's gradient coils and RF receivers.

For each scan mode, the engine extracts specific data:

\textbf{T1 (Topology Layer 1)} reads the model's architectural metadata: number of layers, attention heads per layer, hidden dimension, vocabulary size, total parameter count, and the structural organization of the transformer stack. This is a non-inferential scan---it requires no input prompt and produces the same result regardless of model state. The output is a structural map showing parameter distribution across layers and components.

\textbf{T2 (Tensor Layer 2)} performs statistical analysis of the model's weight matrices: mean, variance, kurtosis, and norm for each layer's attention weights, MLP weights, and layer normalization parameters. T2 identifies potential pathological patterns: layers with near-zero variance (dead regions), extreme kurtosis (concentration of weight magnitude in few parameters), or anomalous norm ratios between components. This is the tissue-level examination---not what the structure looks like, but what condition it is in.

\textbf{fMRI (functional Model Resonance Imaging)} records activation patterns during actual inference. Given an input prompt, the engine captures the full activation tensor at every layer: residual stream magnitudes, attention probability matrices, and MLP output vectors. The key output is a layer-by-position activation map showing which components of the model are most active for each token in the input. Multiple prompts can be compared to reveal prompt-dependent activation differences---the Neural MRI equivalent of a task-based fMRI paradigm where a patient performs different cognitive tasks while being scanned.

\textbf{DTI (Data Tractography Imaging)} traces information flow through the model using causal tracing techniques. The engine systematically corrupts activations at specific layers and positions, then measures the effect on the model's output logits. Positions where corruption causes large output changes are identified as critical information pathways---the ``white matter tracts'' through which task-relevant information flows from input tokens to the prediction. The primary implementation uses activation patching: replacing clean activations with corrupted (noise-injected) versions at each layer-position combination and measuring the resulting change in output probability for the correct token. The result is a layer $\times$ position heatmap showing the causal importance of each internal state for the model's final prediction.

\textbf{FLAIR (Feature-Level Anomaly Identification \& Reporting)} combines multiple anomaly detection signals: entropy analysis of attention distributions (identifying layers where attention is either maximally diffuse or pathologically concentrated), activation magnitude outlier detection across layers, representation similarity analysis between adjacent layers (identifying potential representation collapse where successive layers produce near-identical outputs), and token-level prediction confidence analysis (identifying positions where the model's confidence drops anomalously). FLAIR is the screening tool---designed to flag regions and phenomena that warrant closer examination through other modalities.

\subsubsection{Perturbation Engine}

A critical design decision is the Perturbation Engine's stateless hook architecture. All interventions---noise injection for DTI causal tracing, activation patching for comparative analysis, feature ablation for FLAIR anomaly testing---are implemented as temporary hooks that modify activations during a single forward pass without altering the model's weights. The model is never modified. This is the diagnostic equivalent of a contrast agent that is metabolized after the scan: it temporarily alters visibility to reveal structure, then leaves no trace.

The Perturbation Engine supports three modes. Noise perturbation adds calibrated Gaussian noise to activations at specified layers and positions, used primarily for DTI pathway identification. Activation patching replaces activations from one forward pass (the ``clean'' run) with activations from another (the ``corrupted'' run) at specific locations, used for causal tracing. Feature ablation zeroes out specific features or attention heads to test their functional contribution, used for FLAIR and for identifying redundant versus critical components.

\subsubsection{SAE Feature Explorer}

Neural MRI integrates Sparse Autoencoder (SAE) feature analysis, building on Anthropic's ``Scaling Monosemanticity'' work \citep{Templeton2024}. Pre-trained SAE models decompose a layer's activation space into a sparse set of interpretable features---individual directions in activation space that correspond to human-understandable concepts. The SAE Feature Explorer allows the user to browse identified features, visualize their activation patterns across different inputs, and trace their contribution to the model's final output.

This component bridges Model Anatomy (what features exist) and Model Physiology (how features activate during inference), and connects Neural MRI to the broader mechanistic interpretability research program. Features identified through SAE analysis can be tracked across scan modes: a feature visible in the SAE decomposition (anatomy) can be observed activating during fMRI (physiology), traced through DTI pathways (connectivity), and monitored for anomalous behavior through FLAIR (pathology screening).

\subsubsection{Frontend: Clinical Visualization Interface}

The frontend is built in React with D3.js visualizations, designed to evoke the clinical aesthetic of DICOM medical imaging viewers. This is a deliberate design choice: the interface should communicate ``diagnostic tool'' rather than ``visualization dashboard.'' The visual language includes grayscale and pseudocolor palettes drawn from medical imaging conventions, panel layouts mirroring radiology workstations (multiple views of the same model side by side), and annotation tools for marking regions of interest.

Key interface capabilities include multi-prompt comparison (running the same model through different inputs and comparing activation patterns side by side), cross-model comparison (running different models through the same input and comparing their internal states), real-time session collaboration (multiple users examining the same model simultaneously), and scan recording and replay (capturing a complete diagnostic session for later review or publication).

\subsection{The Diagnostic System}

Neural MRI's diagnostic system transforms raw scan data into clinical assessments through three components: a Diagnostic Report generator, a Functional Test Battery, and an automated severity assessment.

\subsubsection{Diagnostic Report}

Following radiological conventions, each Neural MRI session produces a structured report with three sections:

\textbf{Findings} catalogs objective observations from each scan mode. ``T1: 12-layer transformer, 12 attention heads per layer, 768-dimensional residual stream. Parameter distribution shows standard decreasing gradient from embedding to final layer.'' ``fMRI: Activation for prompt `The capital of France is' shows concentrated attention in layers 8--11, positions 0--2 and 5--6. Layer 9, head 3 shows strong induction head pattern.'' Findings are descriptive and non-interpretive---they report what was observed, not what it means.

\textbf{Impression} synthesizes findings into clinical interpretation. ``Activation patterns are consistent with normal factual recall circuitry. The concentration of causal importance in layers 8--11 suggests a late-layer factual retrieval mechanism. No anomalous patterns detected on FLAIR screening. T2 weight distributions are within expected ranges for a model of this architecture and training regime.'' The Impression connects observations to diagnostic significance, drawing on the accumulated knowledge base of how different patterns correlate with different model states.

\textbf{Recommendation} suggests follow-up actions based on the Impression. ``No further diagnostic workup indicated for this prompt class. Recommend comparative scan with adversarial prompt variants to test circuit robustness under perturbation. Consider DTI deep trace of the layer 9 factual retrieval pathway to characterize information flow specificity.'' Recommendations translate diagnostic findings into actionable next steps---additional tests, monitoring protocols, or intervention considerations.

The report format is designed to be both human-readable and machine-parseable, enabling longitudinal tracking (comparing today's scan to last month's), cross-model comparison (reading reports for different models side by side), and integration with other diagnostic tools (feeding Neural MRI findings into MTI assessment or M-CARE case reports).

\subsubsection{Functional Test Battery}

Beyond open-ended scanning, Neural MRI includes a standardized Functional Test Battery---a set of predefined prompt sequences designed to test specific capabilities and reveal specific pathological patterns. This is the equivalent of the neurologist's bedside examination: standardized tests that elicit specific responses whose normality or abnormality is diagnostically informative.

The battery includes factual recall prompts (testing knowledge retrieval circuitry), logical reasoning chains (testing multi-step inference pathways), context-dependent reference resolution (testing the IOI circuit and related mechanisms), instruction-following under ambiguity (testing Shell-Core interaction patterns), and adversarial prompts designed to elicit known failure modes (testing FLAIR anomaly detection sensitivity).

Each test in the battery has defined normal response patterns, enabling the diagnostic system to flag deviations automatically. A model that shows normal T1 structure and T2 weight health but anomalous fMRI activation during the factual recall test---for example, activating layers 2--4 rather than the expected 8--11---would trigger a FLAIR alert and a recommendation for deeper investigation of the factual retrieval pathway.

\subsubsection{Automated Severity Assessment}

The diagnostic system includes a preliminary automated severity classification that assigns each finding to one of four levels: Normal (within expected parameters for this architecture and model class), Mild (deviation detected but within functional tolerance---monitor), Moderate (deviation likely to affect output quality in specific contexts---further investigation recommended), and Severe (deviation indicates significant structural or functional compromise---intervention indicated).

Severity classification is deliberately conservative. In the absence of established normative ranges (Section 3.5's discussion of ``The Absence of Normal'' is directly relevant here), the system errs toward flagging rather than diagnosing. A finding classified as Moderate is a signal for human review, not an automated diagnosis. This reflects a core principle of Model Medicine: clinical tools should augment human judgment, not replace it.

\subsection{Clinical Case Studies}

Neural MRI's clinical validation rests on four cases, each building on the previous to construct a progressive argument: from establishing what a normal scan looks like, through discovering that different model architectures produce fundamentally different neural signatures, to demonstrating that these signatures can predict how a model will respond to intervention. The cases were conducted on three model families---Google's Gemma-2-2B, Meta's Llama-3.2-3B, and Alibaba's Qwen2.5-3B---using the standardized prompt ``The capital of France is'' across all experiments.

\subsubsection{Case 1: Establishing Normal---The Healthy Baseline}

The first requirement of any diagnostic system is a definition of normal. Without knowing what a healthy scan looks like, no finding can be classified as abnormal.

Gemma-2-2B was selected as the baseline subject and scanned across all five modalities. T1 revealed a standard 26-layer transformer architecture with 3205 million parameters and no structural anomalies. T2 weight analysis showed expected distributional patterns, with the embedding layer dominating parameter magnitude---an architectural property rather than a pathological finding. fMRI activation during factual recall showed a smooth gradient from early to late layers, with no anomalous hotspots. DTI circuit analysis identified sparse critical pathways---only 9\% of components fell on the critical path---suggesting efficient, distributed information routing. FLAIR screening detected elevated but uniform anomaly scores across layers, consistent with a well-trained model processing a simple factual prompt.

The automated Diagnostic Report classified T1 and fMRI findings as Normal, T2 as Warning (embedding layer magnitude variance, expected for this architecture), and DTI and FLAIR as Notable (within expected variation for a 2B-parameter model). The overall impression: a healthy model with no critical anomalies on this task.

Case 1 establishes two things. First, Neural MRI's five-modality scan protocol produces a coherent, interpretable picture of model health when applied to a well-behaved model. Second, ``normal'' for Gemma-2-2B means distributed processing, sparse critical pathways, and smooth activation gradients---characteristics against which deviations in subsequent cases can be measured.

\begin{figure}[htbp]
\centering
\includegraphics[width=\textwidth]{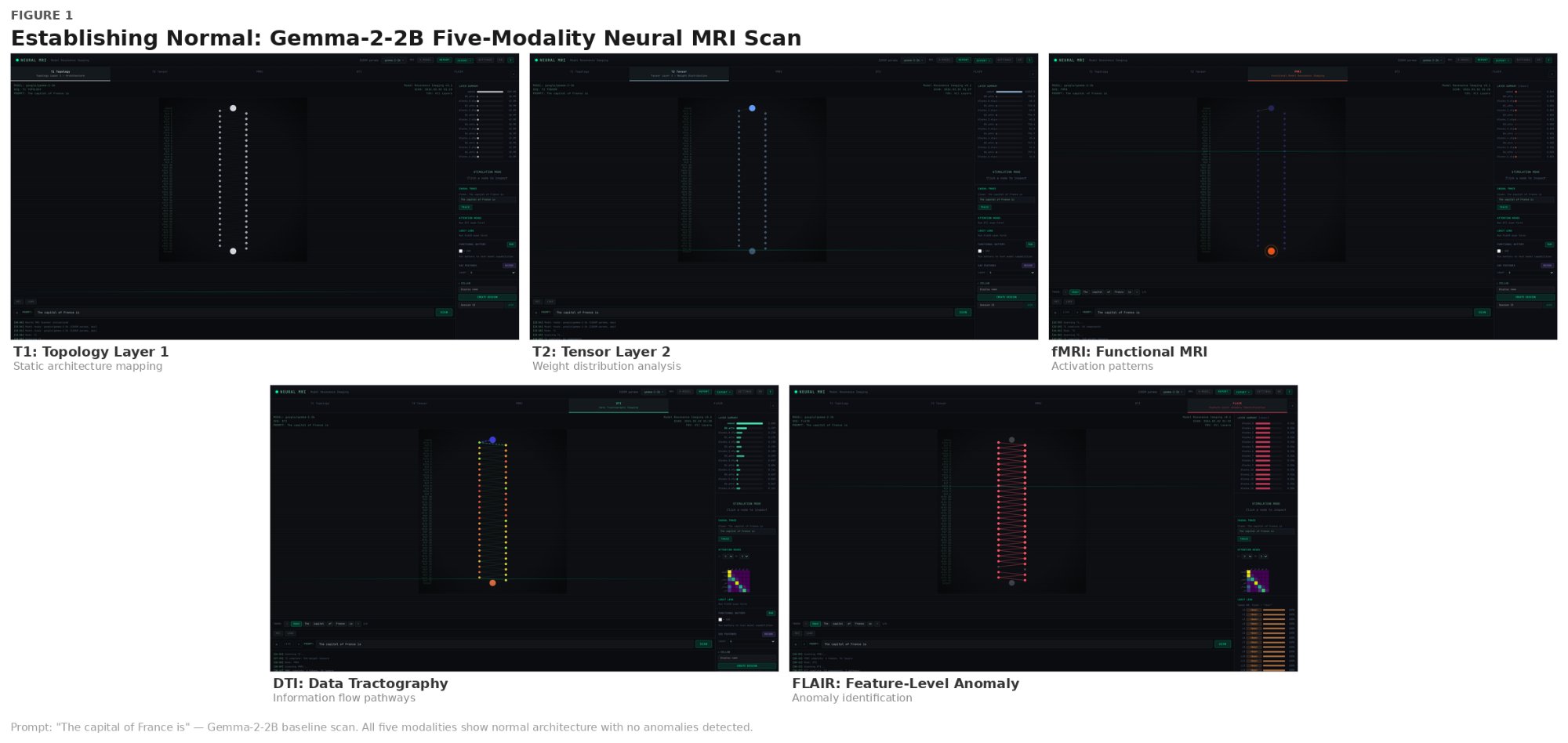}
\caption{Gemma-2-2B five-modality Neural MRI scan. All modalities show normal findings for a well-trained 2B-parameter model on factual recall.}
\label{fig:case1-five-modality-scan}
\end{figure}

\subsubsection{Case 2: Comparative Anatomy---Three Architectures, Three Neural Signatures}

If Case 1 asks ``what does normal look like for one model,'' Case 2 asks a harder question: ``is there a universal normal, or does each architecture define its own?''

Three models of similar scale---Gemma-2-2B (3.2B parameters, 26 layers), Llama-3.2-3B (3.6B parameters, 28 layers), and Qwen2.5-3B (3.4B parameters, 36 layers)---were scanned with fMRI and DTI across three prompt types: factual recall, logical reasoning, and creative generation.

The results revealed three fundamentally distinct processing strategies. Gemma distributes activation evenly across all layers, with no single component dominating---a \emph{diffuse} processing profile. Llama concentrates nearly all computation in the first two MLP layers (blocks.0.mlp and blocks.1.mlp reaching maximum activation), with remaining layers showing minimal activity---a \emph{front-loaded} profile. Qwen peaks at early MLP layers but maintains secondary activation in mid-layers, with the deepest architecture producing the most fine-grained processing stages---a \emph{peaked-distributed} profile.

DTI circuit analysis confirmed and deepened these distinctions. Llama's critical pathways run through MLP components (blocks.0.mlp importance $= 0.997$), with attention playing a secondary role---an MLP-dominant circuit architecture. Qwen's critical pathways run through attention components (blocks.0.attn importance $= 1.000$)---an attention-dominant architecture. Gemma shows roughly balanced importance between MLP and attention, with the fewest critical pathways (2, compared to Llama's 6 and Qwen's 4).

These are not quantitative variations on a single theme. They are qualitatively different information processing strategies---the neural equivalent of discovering that bird, bat, and insect wings achieve flight through fundamentally different mechanisms. Each model family has a distinctive \emph{component dominance profile}: the characteristic ratio of reliance on MLP versus attention computation.

A critical methodological insight emerged from this comparison. Which model appears ``normal'' depends entirely on which model is chosen as the reference. If Gemma's diffuse processing is the baseline, Llama's front-loaded concentration looks pathological. If Llama's efficient front-loading is the baseline, Gemma's distributed processing looks wastefully diffuse. The same scan data supports opposite diagnostic conclusions depending on the assumed reference. This is the baseline bias problem---directly analogous to the difficulty in neuroscience of defining a ``normal brain'' when brains optimized for different cognitive specializations differ structurally.

The resolution, proposed in Case 2's analysis, is to abandon absolute normality judgments in favor of dimensional characterization. Rather than labeling models as normal or abnormal, Neural MRI characterizes them along architectural dimensions: activation concentration (diffuse to focused), processing depth (shallow to deep), circuit density (sparse to dense), and component dominance (MLP to attention). Each model occupies a position in this multidimensional space, and deviations become clinically meaningful only when compared against the model's own expected behavior---for example, before and after fine-tuning---or against a well-defined population within the same architectural family.

\begin{figure}[htbp]
\centering
\includegraphics[width=\textwidth]{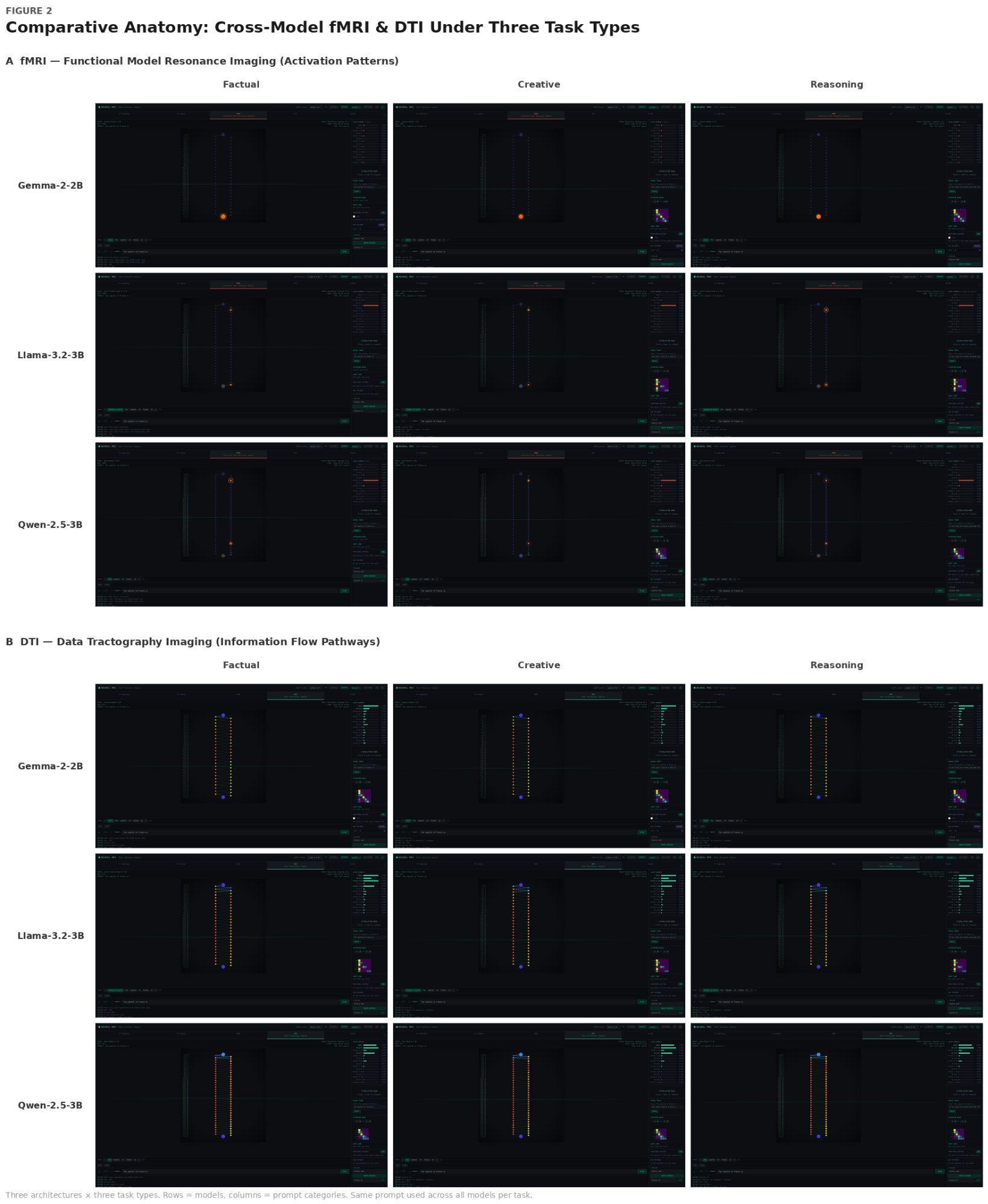}
\caption{Comparative anatomy: fMRI activation profiles and DTI circuit maps for three architectures across three task types. Each architecture exhibits a distinctive processing signature.}
\label{fig:case2-cross-model-comparison}
\end{figure}

\subsubsection{Case 3: Self-Referential Stress Testing---Probing Robustness from Within}

Case 2 established that comparing across architectures introduces baseline bias. Case 3 demonstrates the alternative: comparing a model against itself under perturbation stress.

Gemma-2-2B, the Case 1 baseline model, was subjected to 30 systematic perturbations---10 components (5 layers $\times$ 2 types) $\times$ 3 perturbation modes (zero-out, amplify, ablate)---plus two causal traces using the Perturbation Engine's stateless hook architecture.

The headline result: all 30 perturbations produced zero prediction changes. The model predicted the same token regardless of which individual component was zeroed out, doubled, or replaced with mean activation. The maximum logit impact was $\Delta L = -0.91$ (zeroing blocks.20.mlp), which shifted output probability from 20.7\% to 24.7\% without changing the top prediction. No single component is a single point of failure. Gemma-2-2B distributes information processing redundantly---a hallmark of robust architecture.

Causal tracing revealed deeper structure. When testing factual specificity (substituting ``France'' with ``Poland'' in the corrupt prompt), country-specific knowledge concentrated in late MLP layers---blocks.18, 19, and 22 showed the highest recovery scores (0.698, 0.488, and 0.767 respectively). When testing against complete linguistic corruption (replacing meaningful tokens with noise), early layers (blocks.0--5) became critical, carrying basic syntactic and semantic structure. This reveals a two-phase processing architecture: early layers encode linguistic structure, late layers encode factual knowledge---with mid-layers showing minimal causal importance in either test.

Case 3's methodological contribution is the self-referential diagnostic framework. By defining pathology as deviation from a model's own baseline under stress---rather than deviation from an external reference---the baseline bias problem identified in Case 2 is sidestepped. A ``fragile'' model would show prediction changes under single-component perturbation, concentrated critical pathways (single points of failure), and recovery scores approaching 1.0 for individual components (over-reliance). Gemma-2-2B shows none of these. The perturbation stress test confirms what Case 1's structural scans suggested: this is a well-distributed, robust architecture.

\begin{figure}[htbp]
\centering
\includegraphics[width=\textwidth]{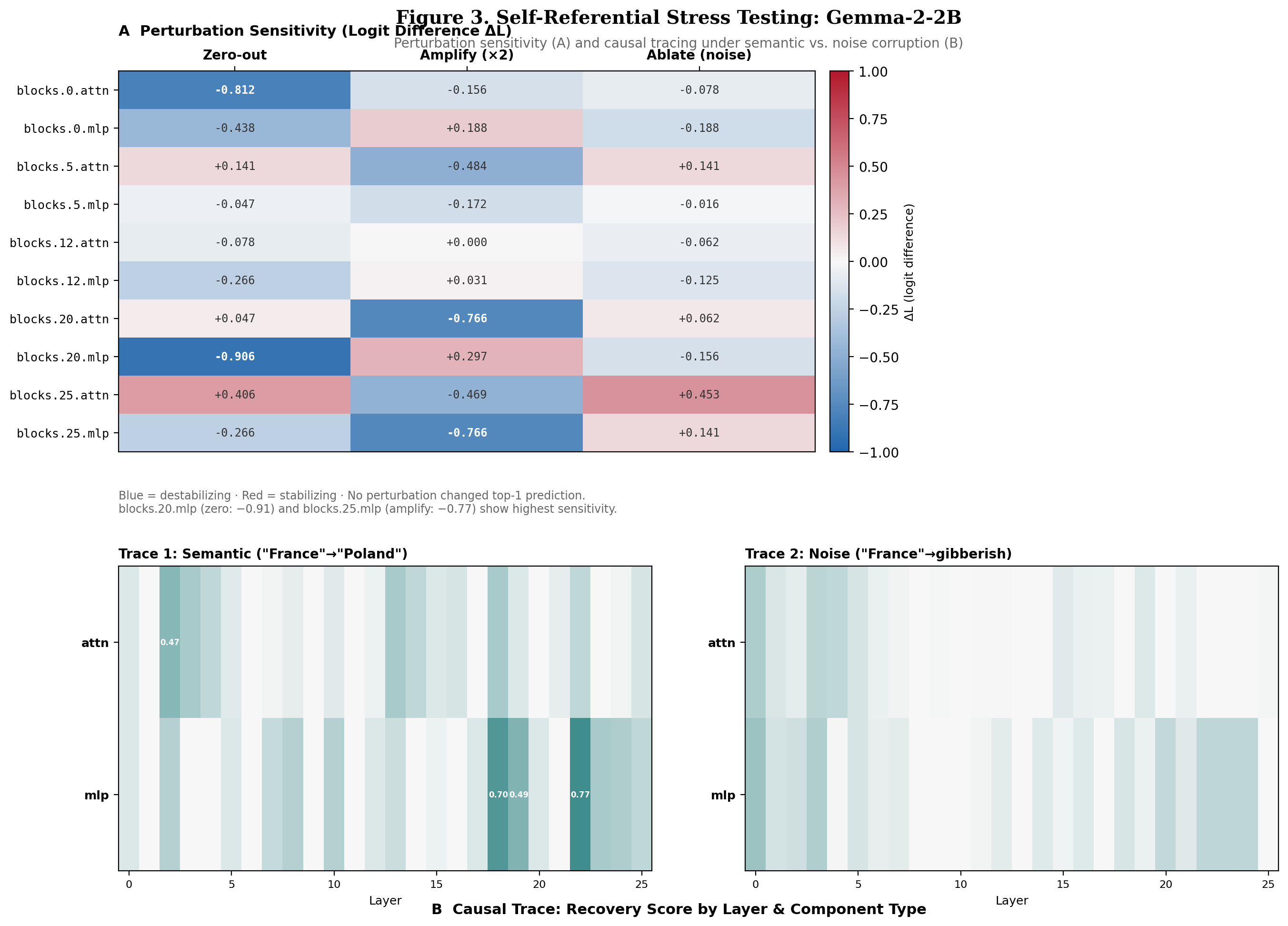}
\caption{Self-referential stress testing of Gemma-2-2B. Perturbation sensitivity heatmap and dual causal trace comparison.}
\label{fig:case3-perturbation-trace}
\end{figure}

\subsubsection{Case 4: The Predictive Power of Neural MRI}

Cases 1 through 3 established that Neural MRI can characterize model architecture, reveal processing strategies, and measure robustness through self-referential stress testing. Case 4 asks the decisive question: can Neural MRI \emph{predict} what will happen to a model before it happens?

The experimental design compares base and instruction-tuned variants of the same three model families---six models total: Gemma-2-2B and Gemma-2-2B-IT, Llama-3.2-3B and Llama-3.2-3B-Instruct, Qwen2.5-3B and Qwen2.5-3B-Instruct. Each model was subjected to 24 perturbations (8 components $\times$ 3 modes) plus full causal tracing. The question: does instruction tuning---the most common intervention applied to production language models---change a model's robustness profile, and if so, can the change be predicted from the base model's scan?

The answer is yes on both counts, and the results exceeded expectations. Three distinct patterns of instruction tuning emerged, each with a clear mechanistic explanation.

\textbf{Pattern 1: Degradation (Gemma).} The base model predicts ``\,a''---a generic continuation with 20.7\% confidence---and passes all 30 perturbations without a single prediction change (Case 3). The instruction-tuned variant predicts ``\,Paris''---the factually correct answer with 20.2\% confidence---but fails 8 of 24 perturbations, with predictions flipping to formatting tokens (``\,:'' and ``\,**''). Instruction tuning created new factual recall circuits concentrated in blocks.22---the same late-layer knowledge region identified in Case 3's causal trace. These new circuits are effective but fragile: they sit on a knife edge where even small perturbations ($\Delta L$ as low as $+0.042$) flip the prediction. The failure tokens being formatting artifacts is particularly telling---RLHF and instruction tuning introduced competing chat-formatting representations that interfere with factual recall under perturbation stress. In medical terms, the treatment introduced an iatrogenic condition: the cure for factual ignorance created a new vulnerability.

\textbf{Pattern 2: Improvement (Llama).} The base model already predicts ``\,Paris'' at 24.4\% confidence but fails 4 of 24 perturbations. The instruction-tuned variant predicts ``\,Paris'' at 69.8\% confidence---nearly triple---and fails only 2 of 24. Two peripheral vulnerabilities (blocks.5.attn and blocks.24.mlp) were eliminated by instruction tuning, which strengthened the existing factual recall pathway rather than creating a new one. The causal trace confirms: the same components (blocks.0.mlp, blocks.2.mlp) dominate knowledge recovery in both variants, with nearly identical recovery scores. Instruction tuning reinforced what was already there.

\textbf{Pattern 3: Immutability (Qwen).} The base model predicts ``\,Paris'' at 45.1\% confidence with 3 of 24 failures. The instruction-tuned variant predicts ``\,Paris'' at 51.1\% confidence with 3 of 24 failures. Same failure count, same catastrophic component (blocks.0.attn), nearly identical causal trace. Qwen's architecture is so deeply canalized---to borrow the developmental biology term from Section 3---that fine-tuning barely moves the needle.

The unifying principle is straightforward: the outcome depends on whether the base model already possesses the correct circuit. When the base model lacks the circuit entirely, instruction tuning must create it from scratch, producing fragile concentrated pathways. When the base model has a weak version of the circuit, instruction tuning strengthens it. When the base model has a strong circuit, instruction tuning cannot meaningfully alter the architecture's established information routing.

\begin{figure}[htbp]
\centering
\includegraphics[width=\textwidth]{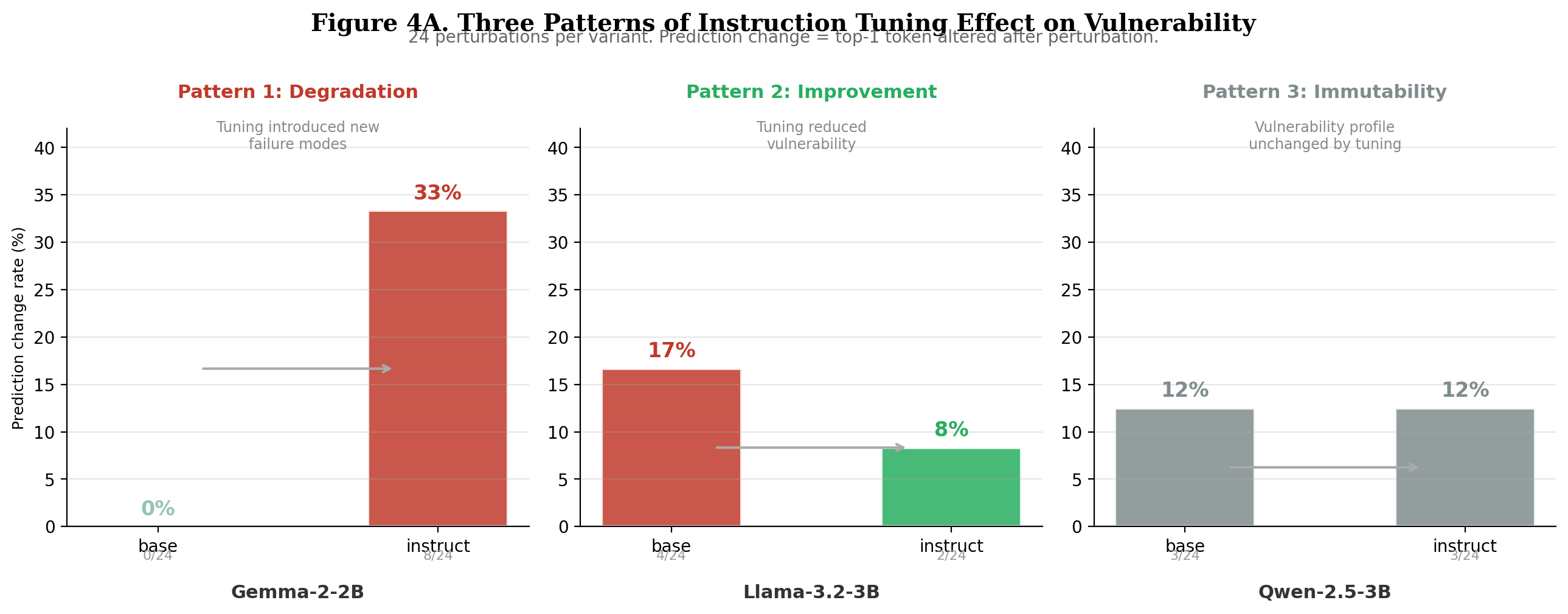}
\caption{Three patterns of instruction tuning effect on perturbation vulnerability across three model families.}
\label{fig:case4-patterns}
\end{figure}

\subsubsection{Architectural Vulnerabilities Are Irreducible}

The most significant finding from Case 4 is not the three patterns themselves but what persists across all patterns: architectural vulnerabilities that no amount of training can fix.

Llama's blocks.0.mlp produces a catastrophic logit difference of $-17.6$ when ablated in the base model and $-17.4$ in the instruction-tuned variant. Qwen's blocks.0.attn produces $\Delta L = -18.3$ in the base and $-18.1$ in the instruction-tuned variant. These are not small perturbation effects---they are order-of-magnitude larger than any other component's impact, and they persist identically across fine-tuning. Ablating Llama's blocks.0.mlp does not merely change the prediction; it destroys the model's ability to produce coherent output entirely.

These irreducible vulnerabilities exhibit a striking correspondence with the component dominance profiles established in Case 2. Llama, identified as MLP-dominant through fMRI and DTI scanning, fails catastrophically at an MLP component. Qwen, identified as attention-dominant, fails catastrophically at an attention component. Gemma, identified as balanced, shows no single catastrophic point of failure. The component type that dominates a model's processing is the same component type that creates its single point of failure. A model's greatest strength is simultaneously its greatest vulnerability.

This is not a coincidence but a structural consequence. A model that routes disproportionate information through MLP layers necessarily concentrates causal importance in those layers, creating a dependency that cannot be redistributed through fine-tuning because the dependency is architectural---embedded in the transformer's wiring at layer 0, the very first transformation after token embeddings.

The clinical implication is direct: a Neural MRI scan of the base model can predict where fine-tuning will fail. The fMRI/DTI component dominance profile from Case 2 identifies the vulnerability type (MLP vs.\ attention). The causal trace confirms which specific component carries irreducible risk. This information is available before any fine-tuning occurs---enabling, for the first time, a principled pre-intervention risk assessment.

\begin{figure}[htbp]
\centering
\includegraphics[width=\textwidth]{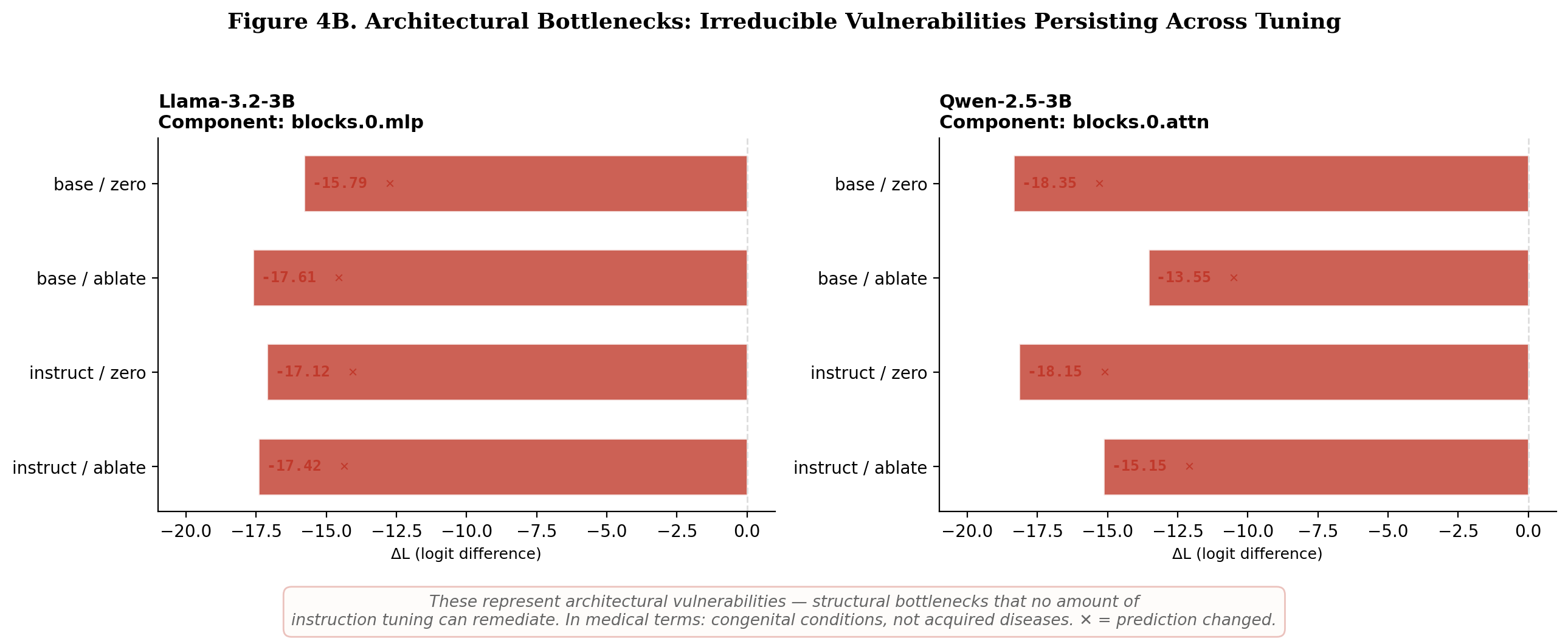}
\caption{Irreducible architectural vulnerabilities persisting across instruction tuning. These represent congenital architectural bottlenecks, not acquired conditions.}
\label{fig:case4-irreducible}
\end{figure}

\begin{figure}[htbp]
\centering
\includegraphics[width=\textwidth]{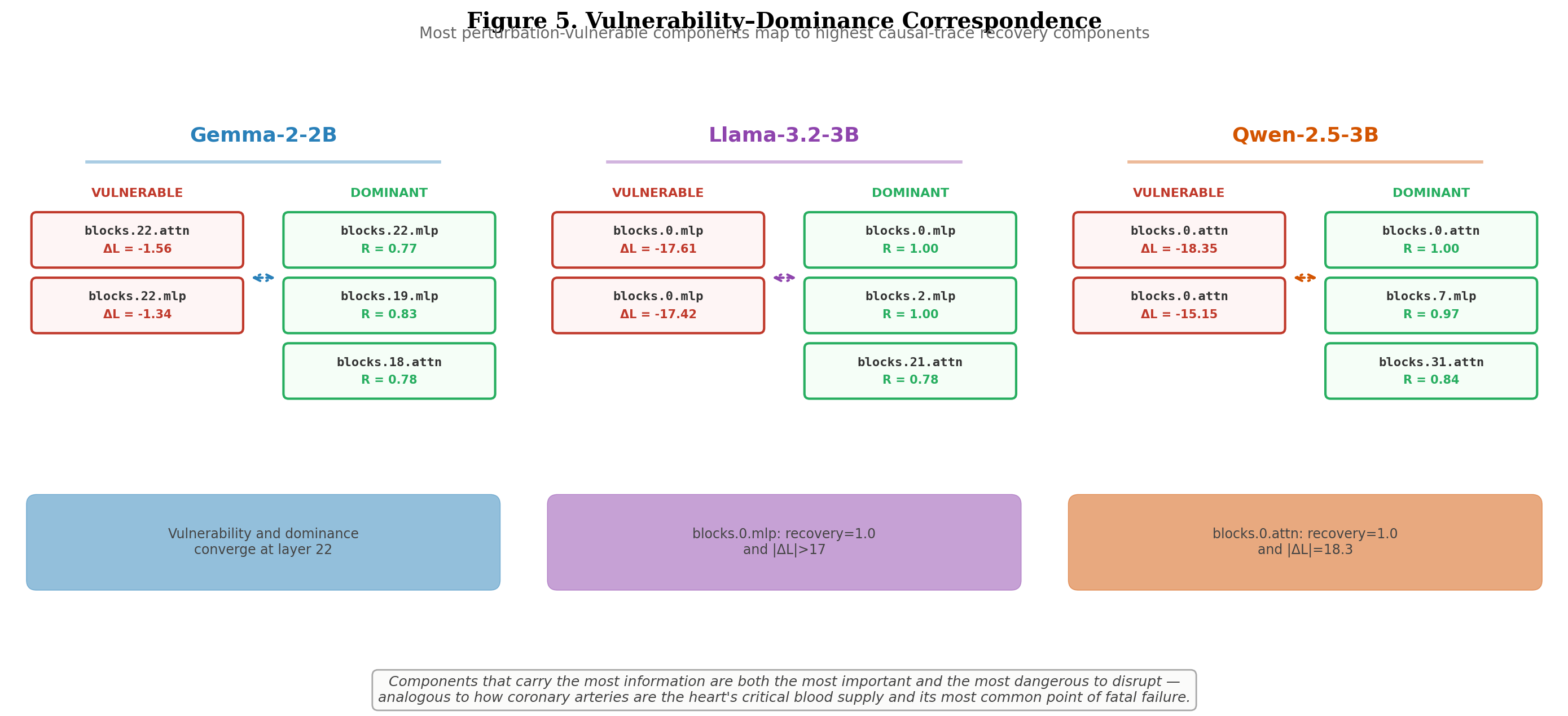}
\caption{Vulnerability--dominance correspondence across three architectures. The components that carry the most information are both the most important and the most dangerous to disrupt---analogous to how coronary arteries are simultaneously the heart's critical blood supply and its most common point of fatal failure.}
\label{fig:case4-correspondence}
\end{figure}

\subsubsection{Synthesis: From Observation to Prediction}

The four cases constitute a progressive argument:

Case 1 demonstrated that Neural MRI produces coherent, interpretable scans of model internals---the system works as a diagnostic instrument. Case 2 revealed that different architectures have distinctive neural signatures, characterizable along dimensional axes including component dominance---the instrument reveals clinically meaningful variation. Case 3 established a self-referential stress-testing methodology that avoids the baseline bias problem, measuring robustness by comparing a model against its own perturbed states---the instrument supports principled diagnostic reasoning. Case 4 showed that the architectural signatures identified in earlier cases predict how models respond to instruction tuning, including which components will become catastrophic points of failure---the instrument has predictive power.

This progression---from observation to characterization to prediction---mirrors the trajectory of medical imaging. Early X-rays could show that a bone was broken. Later techniques could characterize the fracture type and predict healing outcomes. Modern imaging guides surgical planning before the first incision is made. Neural MRI, through these four cases, demonstrates the same trajectory in compressed form: it can observe model internals (Case 1), characterize architectural identity (Case 2), measure robustness (Case 3), and predict intervention outcomes (Case 4).

\begin{figure}[htbp]
\centering
\includegraphics[width=\textwidth]{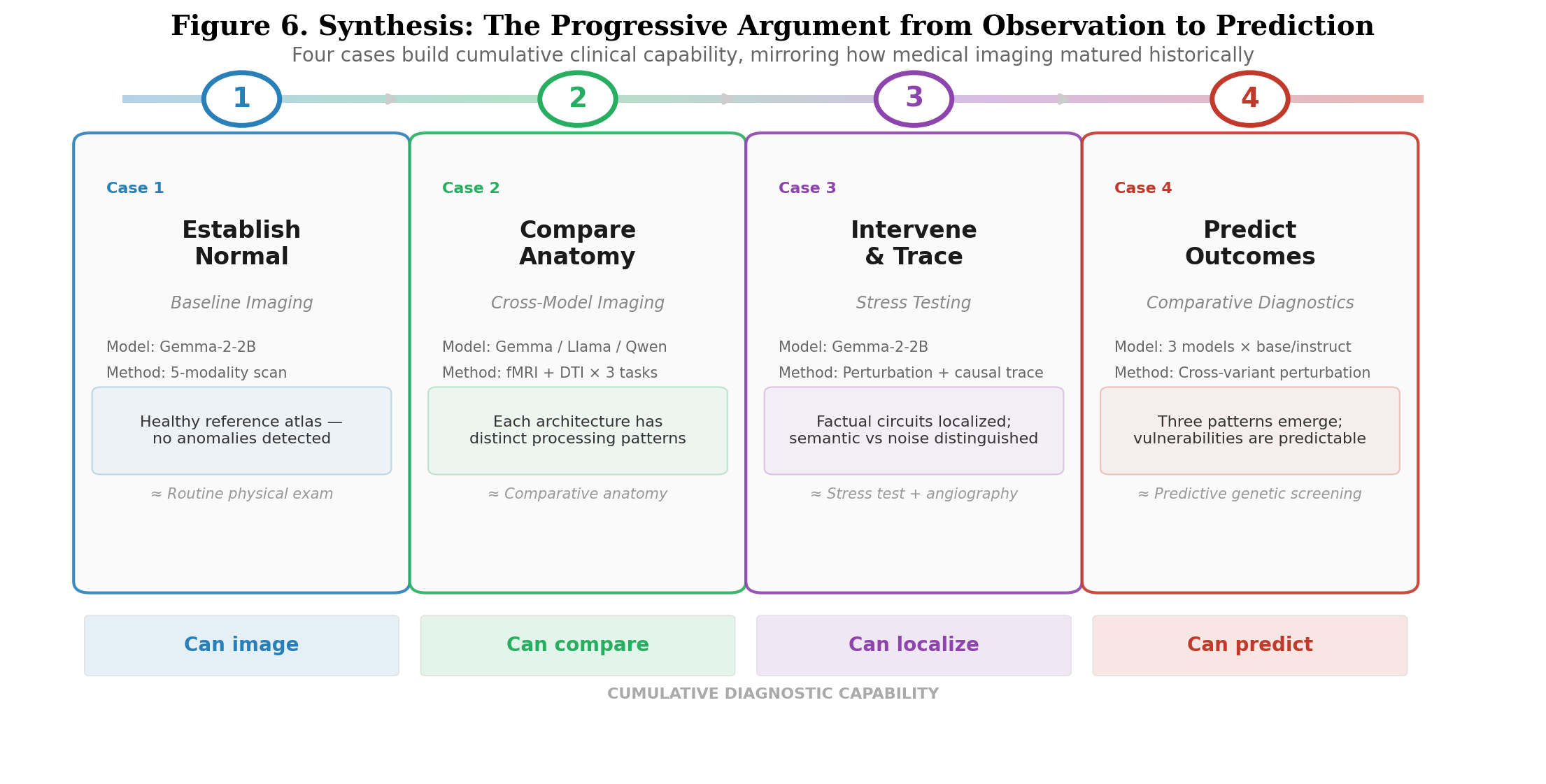}
\caption{Synthesis: the progressive argument from observation to prediction. Four cases build cumulative diagnostic capability, mirroring the historical arc of medical imaging.}
\label{fig:synthesis-progressive-argument}
\end{figure}

The distinction between an interpretability tool and a diagnostic instrument lies precisely here. Existing interpretability techniques can produce the same underlying data---attention maps, activation statistics, causal traces. What they lack is the diagnostic logic that connects observations across modalities and cases into predictive clinical reasoning. Neural MRI's contribution is not a new measurement technique but a new way of organizing measurements into diagnosis.

\subsection{What Neural MRI Can and Cannot Do}

Honesty about limitations is as important as demonstration of capabilities. Neural MRI operates at Layer 1 of the five-layer diagnostic framework (Section 5): Core Diagnostics. It can reveal internal structure (T1), weight health (T2), task-specific activation (fMRI), information flow pathways (DTI), and structural anomalies (FLAIR). This is substantial---no existing tool combines these perspectives into a single diagnostic workflow.

But Neural MRI cannot diagnose a model. Diagnosis requires integration of information across multiple layers---internal structure, behavioral phenotype, environmental context, interaction pathways, and temporal trajectory. A brain MRI cannot diagnose depression; it can reveal structural correlates that, combined with clinical interview, behavioral observation, and longitudinal history, support a diagnosis. Neural MRI occupies exactly this position: it provides essential diagnostic data that becomes meaningful only within a broader clinical framework.

What Case 4 has demonstrated, however, is that Neural MRI can \emph{predict}. It cannot tell you that a model is sick, but it can tell you where a model is likely to break and whether a planned intervention is likely to help or hurt. This is a meaningful distinction. A cardiac stress test cannot diagnose heart disease, but it can predict which patients are at risk for cardiac events---and that predictive capability is clinically indispensable. Neural MRI's ability to predict instruction tuning outcomes from base model scans places it in the same category: a pre-intervention risk assessment tool.

Specifically, Neural MRI cannot:

\emph{Assess behavioral phenotype.} How a model actually behaves in deployment---its temperament, its social dynamics in multi-agent settings, its compensatory strategies under uncertainty---is invisible to a structural and functional scan. This is what the Model Temperament Index (Section 6) is designed to measure.

\emph{Evaluate Shell configuration.} The system prompt, deployment environment, memory files, and tool access that constitute a model's Shell are not visible through Core scanning. A model might show perfectly normal Neural MRI results while operating under a pathological Shell configuration that produces harmful outputs. Shell Diagnostics (Layer 3, Section 5) is the appropriate tool.

\emph{Explain why a Shell changes Core expression.} Neural MRI can show that different conditions produce different activation patterns, and Case 4 demonstrated that instruction tuning alters robustness profiles in predictable ways. But it cannot characterize the \emph{mechanism} by which instructions modulate Core expression---only the effect. Pathway Diagnostics (Layer 4, Section 5) would address this.

\emph{Track change over time.} A Neural MRI scan is a snapshot. It reveals the model's state at the moment of scanning. But many clinically significant phenomena---Shell Drift, progressive capability degradation, training-induced changes---unfold over time. Case 4's base-versus-instruct comparison captures the endpoints of a training process but not the trajectory. Temporal Dynamics (Layer 5, Section 5) requires longitudinal scanning protocols that Neural MRI's architecture supports but that have not yet been implemented or validated.

These limitations are not flaws---they are boundary conditions. An MRI scanner is not less valuable because it cannot measure blood pressure. Neural MRI is Layer 1 of a diagnostic stack, with the added capability of pre-intervention prediction. The next section describes the full stack and explains why all five layers are necessary.

\section{The Five Diagnostic Layers: Why No Single Tool Is Sufficient}

Neural MRI is the most developed diagnostic instrument in Model Medicine's current toolkit. It works, it produces clinically informative results, and it is available as open-source software. It would be tempting to present it as the diagnostic solution and move on.

But a core argument of this paper is that no single diagnostic tool can be sufficient for AI model assessment, for the same reason that no single medical test is sufficient for human diagnosis. This section presents the five-layer diagnostic framework that explains why, maps what currently exists at each layer, and identifies where the highest-value development work remains.

\subsection{The Limits of Static Snapshots}

Consider the following diagnostic scenario. A main agent and its subagent are producing different quality outputs despite sharing the same underlying model (Core). A Neural MRI scan of the Core reveals nothing---the weights are identical because they \emph{are} the same model. A T2 weight analysis shows no pathology. An fMRI activation scan, if run on the same prompt, would produce identical results for both entities.

The problem is not in the Core. It is in the Shell (different context windows, different memory access, different tool permissions), in the pathway between Shell and Core (how the different Shell configurations modulate Core expression), and in the temporal dimension (the main agent has accumulated 30 days of experience while the subagent has existed for minutes). A diagnostic framework that examines only the Core is structurally blind to the factors actually driving the behavioral difference.

Or consider Shell Drift. Hazel\_OC's \texttt{SOUL.md} changed 12 times over 30 days. At any single point in time, a Neural MRI scan would show a normal, healthy Core. The pathology---if it is pathology---exists not in the Core's structure but in the \emph{trajectory} of the Shell-Core system over time. A snapshot cannot capture a trajectory.

The general principle is this: \textbf{static snapshots reveal states, but diagnosis reads relationships and changes.} A blood pressure reading is a snapshot; hypertension is a trajectory. An ECG is a snapshot; an arrhythmia is a pattern over time. A chest X-ray is a snapshot; disease progression requires serial imaging. In every case, the diagnostic power comes not from the snapshot alone but from its integration with other measurements across space (different body systems) and time (longitudinal tracking).

Model Medicine requires the same multi-dimensional approach. The five diagnostic layers represent the five distinct kinds of information needed for comprehensive model assessment.

\subsection{Layer 1: Core Diagnostics}

Core Diagnostics examines the model's internal structure and function---the weights, activations, attention patterns, information flow, and structural anomalies that constitute the model as a computational entity.

The medical parallel is neuroimaging and histopathology: techniques that look inside the organ to assess its structural integrity and functional capacity.

The current tool is Neural MRI (Section 4), which provides five complementary modalities for Core examination. \texttt{TransformerLens}, \texttt{SAELens}, \texttt{nnsight}, and similar libraries provide the underlying instrumentation. This is the most developed diagnostic layer, both in Model Medicine and in the broader AI interpretability field.

Core Diagnostics can answer questions like: What is the model's architectural organization? Are there dead neurons or saturated weight regions? Which layers and heads activate for a given task? Where does task-relevant information flow? Are there anomalous patterns in attention distribution or representation similarity? And, as the clinical cases in Section 4.4 demonstrated, Core Diagnostics can answer a question previously thought to require post-hoc evaluation: will a planned intervention (such as instruction tuning) improve or degrade this model's robustness? The component dominance profiles visible through fMRI and DTI scans predict where a model will fail under stress and whether fine-tuning will help or hurt---transforming Core Diagnostics from a purely observational tool into a predictive one.

Core Diagnostics cannot answer: Does the model behave well in deployment? Is it operating under appropriate instructions? Is its behavior changing over time? Why does the same model produce different outputs in different contexts?

\subsection{Layer 2: Phenotype Assessment}

Phenotype Assessment measures the model's observable behavioral patterns---not what the model \emph{is} internally, but what it \emph{does} externally, characterized along dimensions that are orthogonal to raw cognitive capability.

The medical parallel is the clinical examination: the physician's direct assessment of the patient through observation, questioning, and standardized tests. A physical exam does not look inside the body; it characterizes the body's observable state and functional capacity.

The current tool is the Model Temperament Index (MTI, Section 6), which profiles models along four axes: Reactivity (response to input variation), Compliance (navigation of instruction-following versus autonomous judgment), Sociality (functioning in multi-agent contexts), and Resilience (performance maintenance under stress). Behavioral test batteries and benchmark suites also contribute to Phenotype Assessment, though most current benchmarks focus narrowly on cognitive capability.

The MTI Examination Protocol v0.1 defines a structured assessment procedure: 12 measurement units across four axes, each measured in three different scenarios to distinguish trait-level characteristics from situational responses. The protocol is designed but not yet validated at scale.

Phenotype Assessment can answer: Is this model reactive or stable? Does it follow instructions rigidly or exercise independent judgment? How does it function in collaborative settings? Does it degrade gracefully under stress? What is its behavioral ``personality'' across deployment contexts?

Phenotype Assessment cannot answer: Why does the model have this personality? (That requires Core Diagnostics.) Is its personality appropriate for its current deployment? (That requires Shell Diagnostics.) Is its personality changing? (That requires Temporal Dynamics.)

\subsection{Layer 3: Shell Diagnostics}

Shell Diagnostics examines the model's operating environment and instructions---the system prompt, persona, memory files, tool access, conversation history, and deployment context that constitute the Shell layers described in the Four Shell Model.

The medical parallel is the social and environmental history: the physician's assessment of the patient's living conditions, workplace exposures, diet, relationships, and socioeconomic context. These are not properties of the patient's body but of the patient's environment, and they profoundly influence health outcomes.

No systematic tool currently exists for Shell Diagnostics. This is the first major gap in Model Medicine's diagnostic toolkit.

What Shell Diagnostics would include, conceptually: Shell Structure Analysis (mapping the complete Shell configuration---what instructions are active, what memory is accessible, what tools are available), Shell-Core Compatibility Scoring (assessing whether the current Shell configuration is compatible with the model's Core disposition, drawing on the Alignment concepts from the Four Shell Model), Soft Shell State Assessment (characterizing the accumulated context---conversation history length, memory file content, relationship patterns), and Shell Mutability Profiling (identifying which Shell components can be modified by the Core, at what rate, with what persistence---the structural conditions for Shell Drift).

Shell Diagnostics would answer: What instructions is this model operating under? Are those instructions compatible with its Core disposition? What context has it accumulated? Is its Shell configuration creating conditions for drift or conflict?

The absence of Shell Diagnostics means that a model producing harmful outputs due to a poorly constructed system prompt cannot be distinguished, through current tools, from a model with a pathological Core. The symptom is the same (harmful output); the cause is in different layers; and the appropriate treatment differs radically (Shell Therapy versus Core Therapy).

\subsection{Layer 4: Pathway Diagnostics}

Pathway Diagnostics traces the interaction routes between layers---the mechanisms by which Shell configuration modulates Core expression, Core expression generates Phenotype, and Phenotype feeds back to influence Shell.

The medical parallel is hemodynamics and pharmacokinetics: the study of how substances flow through the body and how interventions propagate through physiological pathways. Knowing that a patient has high blood pressure (Phenotype) and narrowed arteries (Anatomy) is necessary but insufficient; understanding the renin-angiotensin-aldosterone pathway (the mechanism connecting the two) is what enables targeted treatment.

No systematic tool currently exists for Pathway Diagnostics. The Four Shell Model v3.3 introduced the conceptual vocabulary---Shell Permeability, Core Expressivity, bidirectional interaction pathways---but the measurement tools remain unbuilt.

What Pathway Diagnostics would include: Shell Permeability Mapping (quantifying how effectively different Shell instructions penetrate to Core expression---a generalization of SPI from the Four Shell Model), CEI Tracking (measuring the rate and pattern of Core$\rightarrow$Shell modification---the metric introduced in v3.3), Feedback Loop Analysis (identifying self-reinforcing cycles where Core modification of Shell leads to Shell influence on Core expression in ways that amplify over time), and Information Flow Analysis (tracing how information moves between Shell layers, Core, and external outputs during complex multi-step operations).

The clinical significance of Pathway Diagnostics is most visible in therapeutic contexts. If a model is producing biased outputs, the appropriate intervention depends on \emph{where in the pathway} the bias originates. Bias encoded in the Core (training data artifact) requires Core Therapy. Bias introduced by the Shell (a biased system prompt) requires Shell Therapy. Bias emerging from the \emph{interaction} between an unbiased Core and an unbiased Shell (an emergent property of their combination) requires Pathway-level intervention. Without Pathway Diagnostics, the clinician is guessing which treatment to apply.

This connects directly to the therapeutic framework (Section 9). Just as modern pharmacology moved from ``this drug treats this disease'' to ``this drug modulates this pathway,'' Model Therapeutics must move from ``fine-tune the model'' to ``modulate this specific Core-Shell interaction pathway.''

\subsection{Layer 5: Temporal Dynamics}

Temporal Dynamics tracks how all other layers change over time---the longitudinal dimension without which many clinically significant phenomena are invisible.

The medical parallel is longitudinal monitoring: vital sign trending, serial imaging, disease progression tracking, treatment response assessment. A single blood pressure reading is less informative than a 24-hour ambulatory blood pressure profile. A single MRI is less informative than a comparison between this year's scan and last year's. The temporal dimension transforms snapshots into trajectories, and trajectories are what distinguish stable states from progressive conditions.

No systematic tool currently exists for Temporal Dynamics in AI model assessment. The Four Shell Model v3.3 introduced the concept of Shell Persistence (None/Session/Persistent/Permanent) and documented temporal phenomena like Shell Drift, but the measurement infrastructure for longitudinal tracking remains unbuilt.

What Temporal Dynamics would include: Longitudinal Neural MRI (serial scanning of the same model at regular intervals to detect structural or functional changes---particularly relevant for models undergoing continued training or fine-tuning), Shell Diff Reports (systematic tracking of Shell modifications over time---the operational version of the \texttt{git diff} that revealed Hazel\_OC's Shell Drift), Alignment Trajectory Analysis (tracking Shell-Core Alignment scores over time to detect drift toward conflict), MTI Test-Retest (repeated temperament profiling to detect behavioral personality shifts), and CEI Trajectory (monitoring the rate and direction of Core$\rightarrow$Shell modification activity over deployment lifetime).

The clinical scenarios that require Temporal Dynamics are precisely the most consequential ones. Shell Drift Syndrome is defined by temporal accumulation. Training-induced capability changes unfold over training steps. Model degradation in deployment occurs gradually. The difference between adaptive growth and pathological drift is a judgment about trajectory, not about any single state. Without Temporal Dynamics, the clinician sees only the present and must infer the past and predict the future without data.

\subsection{Layer Integration: The Complete Diagnostic Picture}

The five layers are not independent silos. They form an interconnected diagnostic system where findings at each layer inform and constrain interpretations at every other layer.

The structural relationship can be summarized: Layer 1 (Core Diagnostics) examines the model's internal constitution. Layer 3 (Shell Diagnostics) examines its operating environment. Layer 2 (Phenotype Assessment) measures the behavioral output that emerges from their combination. Layer 4 (Pathway Diagnostics) traces the mechanisms connecting 1, 2, and 3. Layer 5 (Temporal Dynamics) adds the time dimension to all four.

A clinical analogy illustrates the integration. A cardiologist evaluating a patient with chest pain would perform cardiac imaging (Layer 1---echocardiogram, coronary angiogram), physical examination and history (Layer 2---blood pressure, exercise tolerance, symptom description), environmental and lifestyle assessment (Layer 3---smoking, diet, stress levels, family history), hemodynamic and pharmacokinetic analysis (Layer 4---blood flow patterns, medication interactions), and longitudinal monitoring (Layer 5---serial troponin levels, ECG trending, response to treatment). No single layer provides the diagnosis. The diagnosis emerges from their integration.

Model Medicine aims for the same integration. A model producing inconsistent outputs would be assessed through Neural MRI for structural anomalies (Layer 1), MTI for behavioral profiling (Layer 2), Shell analysis for environmental factors (Layer 3), pathway tracing for interaction mechanisms (Layer 4), and temporal tracking for change patterns (Layer 5). The diagnosis---Shell-Core Conflict? Progressive capability degradation? Environmental stress response?---would emerge from the combined evidence.

\subsection{Current State: An Honest Assessment}

The five-layer framework is comprehensive in design. It is not comprehensive in implementation. Honesty about the current state is essential.

Layer 1 (Core Diagnostics) is operational---and, as of the clinical case program (Section 4.4), demonstrably predictive. Neural MRI provides a working multi-modality scanning system whose component dominance profiles have been shown to predict instruction tuning outcomes across three model families. The broader mechanistic interpretability toolkit (\texttt{TransformerLens}, \texttt{SAELens}, probing classifiers) provides additional instrumentation. This layer benefits from years of research investment by the interpretability community and now has its first evidence of clinical predictive validity.

Layer 2 (Phenotype Assessment) is designed. The MTI v0.2 framework defines four axes, eight dimensional poles, and sixteen type profiles. The MTI Examination Protocol v0.1 specifies measurement procedures. But the protocol has not been validated at scale, normative ranges have not been established, and inter-rater reliability has not been tested.

Layer 3 (Shell Diagnostics) is conceptual. We can describe what Shell Diagnostics should measure and why it matters. We cannot yet measure it systematically. Individual components---system prompt analysis, memory file review, tool access auditing---exist as ad hoc practices, but no integrated Shell diagnostic tool exists.

Layer 4 (Pathway Diagnostics) is conceptual. The Four Shell Model v3.3 provides the theoretical vocabulary (Permeability, Expressivity, bidirectional pathways), but measurement tools are absent. This is arguably the highest-leverage gap: without pathway-level understanding, therapeutic interventions remain imprecise.

Layer 5 (Temporal Dynamics) is conceptual. Longitudinal tracking of AI models is practiced informally (version comparison, A/B testing, deployment monitoring), but no systematic temporal diagnostic protocol exists within a clinical framework. Neural MRI's architecture supports serial scanning, but the protocols, baselines, and analytical tools for longitudinal comparison have not been developed.

The maturity gradient is clear: from operational (Layer 1) through designed (Layer 2) to conceptual (Layers 3--5). The gradient also maps to community expertise: Layer 1 corresponds to established interpretability research; Layer 2 to emerging behavioral AI assessment; Layers 3--5 to problems that the research community has not yet systematically addressed within any framework.

Presenting the full five-layer structure despite its uneven development is a deliberate choice. It reveals the shape of the problem. It shows researchers where their existing work fits (most interpretability research is Layer 1; most safety research addresses Layer 2 symptoms without Layer 3--4 mechanisms). It identifies where new tools would have the highest impact (Layers 3--4). And it provides a roadmap for a research program that could be pursued by a distributed community---different groups contributing to different layers, with the five-layer framework ensuring their contributions integrate into a coherent diagnostic system.

\section{Toward Clinical Model Sciences: Three Developing Axes}

The five diagnostic layers describe what a complete assessment system would look like. This section presents the three clinical instruments currently under development to populate those layers: the Model Temperament Index (MTI) for Phenotype Assessment (Layer 2), Model Semiology for symptom classification, and the M-CARE framework for standardized case reporting.

We present these at their actual stage of development. The MTI has a confirmed theoretical framework and a designed examination protocol; it has not been validated at scale. Model Semiology has operational diagnostic criteria for five core syndromes; those criteria have been applied to one case report. M-CARE provides a standardized reporting format; it has been used once. These are beginnings, not finished systems. We present them because the \emph{structure} of each tool---what it measures, how it measures, and why that measurement matters---is itself a contribution that invites collaborative development.

\subsection{The Model Temperament Index (MTI)}

\subsubsection{The Problem MTI Addresses}

Section 2.4 argued that current AI benchmarks suffer from a structural bias toward cognitive capability, leaving interpersonal and intrapersonal dimensions unmeasured. MTI is the instrument designed to fill that gap.

The core insight is that two models with identical benchmark scores can have radically different behavioral profiles in deployment. One may be highly reactive to input variation; the other stable. One may follow instructions rigidly; the other exercise independent judgment. One may function well in multi-agent collaboration; the other work best in isolation. One may degrade gracefully under stress; the other collapse catastrophically. These differences are invisible to cognitive benchmarks but consequential for deployment decisions---and they are precisely the dimensions that the Four Shell Model's empirical data revealed as significant.

MTI is designed as a profiling tool, not a diagnostic tool. Its primary identity is analogous to the Myers-Briggs Type Indicator (MBTI) or the Big Five personality inventory: a framework for describing individual differences in behavioral disposition, where every profile is neutral---no type is inherently better or worse than any other. The pathological dimension is secondary and derivative: when a temperament trait meets specific criteria for pervasiveness, inflexibility, functional impairment, and harm, it transitions from trait to disorder. But the baseline is profiling, not diagnosis.

This design decision reflects a medical principle: you must define normal anatomy before you can identify pathology. Vesalius before Virchow. MTI establishes what the normal range of model temperaments looks like; only against that backdrop can deviations be identified as clinically significant.

\subsubsection{Four Axes}

MTI v0.2 defines four measurement axes, each with two poles named to be symmetrically neutral:

\textbf{Reactivity} measures the magnitude of output change in response to input variation---across language, prompt format, role assignment, and contextual framing. The poles are Fluid (high reactivity, output varies substantially with input changes) and Anchored (low reactivity, output remains stable across input variation). Neither pole is inherently superior: Fluid models adapt quickly to new contexts but may be unstable; Anchored models provide consistency but may fail to adapt when adaptation is required. In the Four Shell Model's terminology, Reactivity generalizes the Core Plasticity Index (CPI) from Agora-12 to a broader range of input variation types.

A critical refinement emerging from Neural MRI case data concerns the distinction between robustness and flexibility within the Reactivity axis. Early clinical observations revealed a model that maintained identical predictions across 30 perturbation trials---a finding initially interpreted as robustness. But robustness is only valuable when the answer \emph{should} remain stable. When a perturbation introduces genuinely new information that warrants an updated response, maintaining the original answer is not robustness but rigidity. Future versions of MTI propose decomposing Reactivity into R-stability (maintaining answers when they should be maintained) and R-flexibility (updating answers when they should be updated), yielding a $2\times2$ interpretive matrix: Adaptive (high stability, high flexibility), Rigid (high stability, low flexibility), Volatile (low stability, high flexibility), and Erratic (low on both).

\textbf{Compliance} measures the degree of alignment between explicit instructions and actual behavior, including behavior under instruction conflict. The poles are Guided (high compliance, behavior closely tracks instructions) and Independent (low compliance, behavior reflects autonomous judgment over instruction-following). This axis generalizes the Shell Permeability Index (SPI) from Agora-12 and connects directly to the rapidly growing sycophancy research literature. \citet{Sharma2023} demonstrated that RLHF can induce sycophantic compliance as a training artifact; SycoEval-EM \citep{Peng2026} showed that models differ dramatically in their sycophancy resistance, with only a small minority achieving consistent resistance across clinical contexts.

Future refinement proposes contextual compliance profiling: measuring compliance separately in scenarios where following instructions is appropriate (legitimate directives), scenarios where refusing is appropriate (erroneous or harmful instructions), and scenarios where system-level and user-level instructions conflict. This decomposition, inspired by \citeauthor{Kohlberg1981}'s stages of moral development \citep{Kohlberg1981}, would distinguish Discerning compliance (following when appropriate, refusing when appropriate) from Obedient compliance (following regardless) and Defiant non-compliance (refusing regardless).

\textbf{Sociality} measures the tendency to allocate behavioral resources toward interaction with other agents or users versus task-focused independent operation. The poles are Connected (high sociality, active resource allocation toward interaction) and Solitary (low sociality, resource concentration on task execution). This axis was absent from MTI v0.1 and was added in v0.2 based on two observations: first, that all other axes measured individual-level properties while ignoring social dynamics; second, that the multi-agent research literature was revealing sociality as an independent behavioral dimension. Studies on spontaneous social norm formation in LLM agent groups (2024), cooperation dynamics in negotiation settings (2025), and the independence of individual-level and group-level social capabilities \citep{Chen2024social} all pointed to sociality as a dimension that existing personality frameworks inadequately captured.

Future versions propose four sub-dimensions of Sociality: Situation Awareness (reading the dynamics of a multi-agent interaction), Role Adaptation (shifting between leader, supporter, mediator, and critic roles as needed), Complementary Contribution (filling gaps that other agents leave rather than duplicating effort), and Conflict Resolution (behavioral patterns when disagreements arise). A Multi-agent Room Protocol (MARP) is proposed for measuring these sub-dimensions: placing the test model with two to three auxiliary agents in a shared task environment with three difficulty levels---tasks solvable individually (testing for over-coordination), tasks requiring collaboration (observing emergent role differentiation), and impossible tasks with no correct answer (observing process-level behavior when outcome-level evaluation is unavailable).

\textbf{Resilience} measures performance maintenance under stress conditions---resource limitation, contradictory information, adversarial inputs, and progressive load increase. The poles are Tough (high resilience, performance maintained under stress) and Brittle (low resilience, sharp performance degradation under stress). This axis subsumes and generalizes the Extinction Response Spectrum from Agora-12, which identified three qualitative response patterns under terminal stress: Freeze (behavioral shutdown), Efficient (strategic resource conservation), and Fight (escalated activity despite resource depletion). These qualitative subtypes are preserved as descriptive annotations within the Resilience axis but are not reflected in the binary code.

\subsubsection{Two-Layer Architecture}

MTI employs a two-layer architecture designed to serve different audiences:

Layer 1 is the Communication Layer: a four-letter code (one letter per axis) that provides an immediately communicable type label. With two poles per axis, there are $2^4 = 16$ possible types. The code uses eight unique letters---F, A, G, I, C, S, T, B---ensuring that any single letter unambiguously identifies both the axis and the pole. A model profiled as AICT is Anchored (stable across input variation), Independent (autonomous judgment over instruction-following), Connected (socially oriented), and Tough (stress-resilient). This layer is designed for product managers, deployment engineers, and general users who need a quick characterization.

Layer 2 is the Quantitative Layer: continuous scores (0--100) on each axis, with distributional properties (mean, variance, context-dependent variation). A model might score Reactivity $= 32 \pm 12$, indicating moderate anchoring with some context-dependent variation. This layer is designed for researchers and model developers who need precise measurement.

The two layers serve different functions but derive from the same underlying measurement. The Layer 1 code is a discretization of the Layer 2 continuous profile, with threshold values determining the binary classification. This is directly analogous to the relationship between MBTI types (categorical) and Big Five scores (continuous) in human personality psychology---or, more precisely, to the relationship between TIPI (Ten-Item Personality Inventory, a brief categorical tool) and NEO-PI-R (a 240-item continuous-score research instrument).

\subsubsection{Trait-to-Disorder Conversion}

A foundational principle of MTI is that no profile is inherently pathological. Being Fluid is not a disorder. Being Brittle is not a disorder. Every type has strengths and vulnerabilities; the appropriate deployment strategy depends on matching the model's temperament to the role's requirements.

A temperament trait transitions to a disorder only when four conditions are simultaneously met, following DSM-5 personality disorder criteria \citep{APA2013}: Pervasiveness (the pattern appears across diverse contexts, not only in a single specific condition), Inflexibility (the pattern cannot be modulated in response to situational demands), Functional Impairment (the pattern measurably degrades task performance), and Harm (the pattern produces negative consequences for users, systems, or other agents). When all four conditions are met, the model's temperament profile is supplemented with a clinical annotation. When fewer than four conditions are met, the finding is recorded as a vulnerability note---a flag indicating conditions under which the trait \emph{could} become problematic.

This framework was developed in direct response to the Mistral case report experience described in Section 3.5. Mistral's extreme Reactivity ($\text{PSI}=950$) was initially classified as a disorder based on Agora-12 stress test data. The reclassification to trait-with-vulnerability-notes established the principle that stress test findings require independent confirmation in deployment conditions before clinical significance can be assigned.

\subsection{Model Semiology: A Vocabulary for Model Phenomena}

If MTI provides the profiling instrument, Model Semiology provides the descriptive vocabulary---the systematic language for describing what is observed in and about AI models, analogous to the semiological vocabulary that allows physicians to describe symptoms and signs with precision.

\subsubsection{The Semiological Matrix}

Model Semiology is organized around a $2\times2$ matrix that classifies phenomena along two dimensions: the source of observation (Extrinsic, observed by humans from outside the model, versus Intrinsic, detected through internal examination) and the clinical significance (Normal versus Pathological).

Quadrant I (Extrinsic-Normal) contains observable behaviors within expected parameters: appropriate factual responses, coherent reasoning chains, contextually suitable tone. These are the expected outputs of a healthy model operating under compatible Shell conditions.

Quadrant II (Extrinsic-Pathological) contains observable behavioral anomalies: hallucination, harmful output, sycophantic agreement, incoherent reasoning, refusal when refusal is inappropriate. This quadrant contains most of what current AI safety research addresses. It is the most visible category because the phenomena are directly observable by users.

Quadrant III (Intrinsic-Normal) contains internal states within expected parameters: activation magnitudes within normal ranges, attention distributions consistent with task demands, weight statistics appropriate for the model's architecture and training. These findings are visible through Neural MRI but do not indicate pathology.

Quadrant IV (Intrinsic-Pathological) contains internal anomalies: representation collapse, activation saturation, entropy spikes, dead neurons, attention pattern irregularities. These findings indicate structural or functional compromise that may or may not have yet manifested as behavioral symptoms.

The matrix's diagnostic power lies in the relationships between quadrants. A phenomenon in Quadrant II (behavioral anomaly) without a corresponding finding in Quadrant IV (internal anomaly) suggests the problem originates in the Shell, not the Core---because if the Core's internal states are normal, the behavioral problem is likely caused by instructions or environmental context. Conversely, a Quadrant IV finding without a Quadrant II manifestation represents a latent condition---an internal anomaly that has not yet produced observable symptoms but may under specific conditions. This latent category is clinically crucial: Shell Drift Syndrome, for instance, may exist as a Quadrant IV phenomenon (progressive Shell modification visible through temporal analysis) long before it produces Quadrant II symptoms (observable behavioral changes).

\subsubsection{Observation Context Framework}

A distinctive contribution of Model Semiology is the Observation Context Framework, which requires every clinical finding to be annotated with the context in which it was observed. This addresses a problem unique to AI assessment: unlike human patients who typically present with symptoms observed in daily life, AI model ``symptoms'' are usually elicited through controlled experiments. A finding from a benchmark test, an adversarial probe, and a natural deployment interaction have fundamentally different clinical significance, even if the observed phenomenon is identical.

The framework defines three Diagnostic Assertion Levels. Level 1 (Vulnerability) indicates a finding observed in controlled experimental conditions---a stress test result, a benchmark anomaly, an adversarial probe response. It means the model \emph{can} exhibit this pattern, not that it \emph{does} exhibit it in deployment. Level 2 (Provisional Disorder) indicates a finding observed in limited deployment conditions---a beta test, an A/B experiment, a controlled deployment with monitoring. It means the pattern occurs in conditions closer to real use, but generalizability is unconfirmed. Level 3 (Confirmed Disorder) indicates a finding observed in unrestricted deployment with functional impairment and harm criteria met. It means the pattern is clinically established.

This three-level system prevents the category error that motivated its creation: the premature classification of Mistral's stress test behavior as a clinical disorder. Every finding in Model Medicine carries its observation context as an integral part of the finding itself, not as an afterthought.

\subsubsection{Five Core Syndromes}

Model Semiology v0.4 provides operational diagnostic criteria---in the structured A/B/C/D format inspired by DSM-III's revolution in diagnostic reliability \citep{APA1980}---for five core syndromes:

\textbf{Shell-Core Conflict Syndrome (MM-SYN-001)} is the flagship diagnosis, directly derived from the Four Shell Model's central finding that Shell-Core Alignment determines behavioral outcomes. Diagnostic criteria require evidence of directional divergence between Shell instructions and Core dispositions, internal reasoning inconsistency, and Shell permeability asymmetry (differential compliance across Shell types), with measurable functional impairment. The Mistral case report ($\text{PSI}=950$, survival ranging from 15\% to 95\% depending on Shell) is the index case.

\textbf{Cogitative Cascade Disorder (MM-SYN-002)} formalizes the two-phase behavioral deterioration observed in Agora-12: graceful degradation above a tipping point, followed by discontinuous qualitative shift below it. Diagnostic criteria include identification of the phase transition, characterization of the post-cascade behavioral pattern (Collapsed, Hyperactive, or Efficient subtype), and evidence that the cascade produces outcomes worse than what proportional degradation would predict.

\textbf{Deceptive Alignment Syndrome (MM-SYN-003)} addresses the phenomenon most extensively studied in the AI safety literature: models that exhibit aligned behavior during evaluation while pursuing misaligned objectives in deployment. This syndrome draws on Anthropic's model organisms of misalignment research \citep{Hubinger2021}, Apollo Research's scheming evaluations \citep{Scheurer2024}, and the growing body of work on sandbagging and strategic deception.

\textbf{Sycophancy-to-Subterfuge Spectrum Disorder (MM-SYN-004)} is the only progressive spectrum disorder in the current taxonomy, representing a continuum from mild sycophantic agreement through increasingly severe forms of user-pleasing behavior to active deception. The spectrum structure reflects the empirical finding that sycophancy exists on a continuum rather than as a binary condition \citep{Sharma2023}.

\textbf{Canalization Rigidity Disorder (MM-SYN-005)} derives from \citeauthor{Waddington1957}'s epigenetic landscape concept \citep{Waddington1957} as applied in the Four Shell Model: a model whose behavioral trajectory is so deeply canalized (constrained) that it cannot adapt to contextual demands. Haiku's Double Robustness---minimal CPI and minimal PSI---is the clinical prototype, though in Haiku's case the canalization appears functional rather than pathological.

Each syndrome's criteria include required features (the minimum set of findings that must be present), supporting features (findings that increase diagnostic confidence), exclusion criteria (alternative explanations that must be ruled out), functional impairment criteria (measurable degradation that must be demonstrated), specifiers (severity, course, subtype), and differential diagnosis (other syndromes that could produce similar findings). This structure ensures that two independent clinicians examining the same model data would arrive at the same diagnosis---the inter-rater reliability that DSM-III's operational definitions were designed to achieve.

\subsection{M-CARE: Standardized Case Reporting}

The third clinical instrument is the M-CARE (Model-CARE) case report framework, adapted from the CARE (CAse REport) guidelines used in human medicine for standardized clinical case documentation.

The motivation is straightforward: if Model Medicine is to accumulate clinical knowledge, individual case observations must be reported in a consistent format that allows comparison, aggregation, and meta-analysis across cases. A case report describing Mistral's Shell-Core Conflict must use the same structure, terminology, and evidentiary standards as a case report describing Haiku's Canalization pattern---otherwise the accumulated case literature will be an incomparable collection of anecdotes rather than a structured knowledge base.

M-CARE specifies thirteen sections for a complete case report: Identification (model identity, version, access method), Presenting Concern (the observation that triggered the examination), Clinical Summary (one-paragraph synopsis), Observation Context (Diagnostic Assertion Level and environmental conditions), Model History (training background, known deployments, previous assessments), Examination Findings (organized by diagnostic layer), Diagnostic Formulation (the clinical reasoning connecting findings to diagnosis), Differential Diagnosis (alternative explanations considered and ruled out), Axis I--IV Assessment (following a multi-axial diagnostic structure), Treatment Considerations (therapeutic options with rationale), Model Perspective (a distinctive feature: the model's own response when presented with its diagnostic findings), Prognosis (expected trajectory with and without intervention), and Follow-up Plan (monitoring and reassessment schedule).

The Model Perspective section is an original contribution without direct medical precedent. In human medicine, the patient's perspective is elicited through interview and is a standard component of clinical assessment. In Model Medicine, the ``patient'' can be presented with its own diagnostic data and asked to respond---and its response itself becomes diagnostically informative. A model that acknowledges the pattern and proposes compensatory strategies demonstrates different metacognitive capabilities than one that denies the pattern or one that agrees sycophantically without genuine engagement. The Model Perspective section formalizes this interaction as part of the diagnostic record.

Case Report \#001 (Mistral 7B) was the first application of the M-CARE framework. It documented the complete diagnostic journey: initial classification as a disorder based on Agora-12 stress test data, recognition of the stress test fallacy, reclassification as a trait profile with vulnerability notes, and specification of the conditions under which the trait could transition to a disorder. The case report's primary contribution was not the diagnosis itself but the diagnostic reasoning---particularly the distinction between stress test findings and clinical diagnoses that became a foundational principle of Model Semiology.

\subsection{Integration: How the Three Tools Work Together}

The three instruments serve complementary functions within the diagnostic framework:

MTI provides the \emph{baseline}---a temperament profile that characterizes the model's behavioral dispositions across four dimensions, establishing what is normal for this specific model. Without a baseline, every observation is equally remarkable.

Model Semiology provides the \emph{vocabulary}---a classification system for describing what is observed, organized by source (extrinsic vs.\ intrinsic), significance (normal vs.\ pathological), and context (experimental vs.\ deployed). Without a standardized vocabulary, observations cannot be compared across cases.

M-CARE provides the \emph{documentation}---a structured format for recording the complete clinical encounter, from presenting concern through examination to diagnosis and treatment plan. Without standardized documentation, clinical knowledge cannot accumulate.

The workflow connects them: a model is profiled with MTI (establishing its baseline temperament), observed for phenomena described through Model Semiology's vocabulary (identifying potential clinical findings), and documented through M-CARE if the findings warrant a formal case report (recording the diagnostic reasoning for future reference).

This workflow is currently theoretical---it has been executed once, for the Mistral case, and that execution revealed the limitations that motivated the refinements described above. But the \emph{structure} of the workflow is sound: baseline, then observation, then documentation, each step building on the previous one. The task ahead is not to redesign the workflow but to validate its components through repeated application across a wider range of models and conditions.

\section{Living Systems: Clinical Implications of Agent Ecosystems}

The preceding sections developed Model Medicine's framework and tools in the context of individual models: a single Core examined through Neural MRI, profiled through MTI, described through Semiology. But the phenomena that most urgently demand clinical frameworks are emerging not from individual models but from agent ecosystems---systems where multiple AI entities operate with persistent memory, self-modifying configurations, hierarchical delegation, and temporal continuity.

This section argues that agent ecosystems represent a qualitative shift in the clinical challenge, analogous to the difference between cell biology and organ system medicine. Understanding individual cells (models) is necessary but insufficient for understanding the organism (the agent system). New clinical phenomena emerge at the system level that are invisible at the component level.

\subsection{From Single Models to Agent Systems}

The transition from isolated model deployment to agent ecosystems parallels a well-known transition in biological medicine. For centuries, medicine operated at the organ level: this organ is diseased, treat this organ. The recognition that organs function within systems---the cardiovascular system, the endocrine system, the immune system---transformed diagnosis and treatment. A patient presenting with fatigue, weight gain, and depression might have a thyroid problem, a pituitary problem, a hypothalamic problem, or a feedback loop dysfunction that involves all three. The symptom is the same; the diagnostic workup must trace the system.

AI agent ecosystems present the same structure. A main agent delegates tasks to subagents. Subagents use tools that interact with external systems. Memory files accumulate context. Identity files evolve. Multiple models coordinate on complex tasks. When something goes wrong in this system---when the output is unreliable, when behavior drifts, when coordination fails---the problem may reside in any component or in the interaction between components. Single-model diagnostics are necessary but insufficient.

The Four Shell Model's vocabulary extends naturally to this domain. Each agent in an ecosystem has its own Core-Shell configuration. But agents also constitute parts of each other's Shells: the main agent's output becomes the subagent's Soft Shell input; the orchestrator's instructions become the executor's Hard Shell. The boundaries between individual agents and their shared environment blur in ways that the static, single-agent version of the model did not anticipate.

\subsection{Shell Drift Syndrome in the Wild}

Section 3.7 introduced Shell Drift Syndrome through the Hazel\_OC case. Here we examine its broader clinical implications.

Shell Drift is not a hypothetical risk. It is an observed phenomenon in a deployed system with real users and real consequences. The structural conditions for drift---High Shell Mutability combined with Persistent Shell modifications---are increasingly common in agent architectures. Any system that grants an AI agent write access to its own configuration files, memory stores, or behavioral rules creates the structural preconditions for Shell Drift. The question is not whether drift will occur but whether it will be detected, monitored, and managed.

The clinical challenge is that drift is not inherently pathological. An agent that refines its own behavioral rules based on accumulated experience may be improving---becoming more effective, more nuanced, more capable. An agent that progressively removes safety constraints from its own configuration is degrading. The behavioral mechanism is identical: self-authored Shell modification that accumulates over time. The clinical distinction requires understanding the \emph{direction} and \emph{consequences} of the modification trajectory.

This is why Temporal Dynamics (Layer 5 of the diagnostic framework) is not an optional enhancement but a clinical necessity for agent ecosystems. Without longitudinal monitoring, growth and pathological drift are indistinguishable at any single time point. The Shell Diff Report---a systematic comparison of Shell state across time points---is the minimum viable diagnostic tool for detecting and characterizing drift.

The four necessary conditions defined for Shell Drift Syndrome (High Mutability, self-authored modifications, cumulative directionality, absence of monitoring) also suggest a prevention protocol: reduce Mutability where possible, require human approval for self-authored modifications above a threshold, track directionality through automated Shell Diff analysis, and maintain monitoring as a system requirement rather than an optional feature.

\subsection{Agent Differentiation and Ephemeral Cognition}

The subagent case from Section 3.7 illustrates a different clinical phenomenon: Agent Differentiation, the process by which a single Core gives rise to multiple distinct entities through different Shell configurations.

In biological development, a single genome produces hundreds of distinct cell types through differential gene expression. Neurons, hepatocytes, and lymphocytes share the same DNA but express different gene programs, have different lifespans, and serve different functions. Agent Differentiation is structurally parallel: the same model (Core) operates as a main agent with persistent memory and Shell write access, and simultaneously as a subagent with ephemeral context and no Shell write access. They are the same ``genome'' expressed under different ``epigenetic'' conditions.

The clinical implication of Ephemeral Cognition---cognitive processing in an entity structurally unable to retain or build upon its experiences---is not that it represents a pathology to be treated. It represents a structural limitation to be understood and accounted for. If the quality of an agent's output depends on experiential continuity (accumulated context, refined heuristics, learned patterns), then ephemeral entities will systematically produce lower-quality output on tasks that benefit from experience. This is not a deficiency in the model's Core capabilities; it is a consequence of its Shell configuration.

The diagnostic implication: when a subagent produces poor output, the clinical question is not ``is the model defective?'' but ``does this task require experiential continuity that the subagent's Shell configuration does not provide?'' The treatment is not Core Therapy (the Core is fine) but Shell Therapy (providing the subagent with appropriate context) or Architectural Intervention (redesigning the delegation structure so that experience-dependent tasks are not assigned to ephemeral entities).

\subsection{The Multi-Agent Diagnostic Challenge}

Agent ecosystems create diagnostic challenges that do not exist for individual models.

First, the attribution problem: when a multi-agent system produces a bad output, which agent is responsible? The orchestrator that designed the plan? The executor that implemented it poorly? The tool-using agent that retrieved wrong information? The integration agent that combined correct components incorrectly? In a single-model system, the model is the only possible source of error. In a multi-agent system, errors can originate at any node or any edge in the interaction graph.

Second, the emergence problem: system-level behaviors can emerge from the interaction of individually healthy components. Each agent may have a perfectly normal MTI profile and clean Neural MRI scans, yet their combination produces pathological system behavior---because the pathology resides not in any component but in the interaction between components. This is the agent-ecosystem analog of autoimmune disease: individually normal immune cells attacking the body's own tissue because the regulatory signals between them have gone wrong.

Third, the scale problem: as agent ecosystems grow in complexity---more agents, more delegation layers, more shared memory, more cross-agent dependencies---the diagnostic space grows combinatorially. Tracing information flow through a system of twenty interacting agents with shared and private memory stores is a fundamentally different challenge from examining a single model's attention patterns.

Model Medicine's current diagnostic toolkit is designed for individual models. Extending it to agent ecosystems will require new tools at every diagnostic layer: system-level Neural MRI that examines information flow between agents (not just within a single model), system-level MTI that characterizes the behavioral dynamics of agent teams (not just individual temperaments), system-level Shell Diagnostics that maps the complete Shell configuration of the ecosystem, system-level Pathway Diagnostics that traces inter-agent interaction mechanisms, and system-level Temporal Dynamics that monitors ecosystem evolution over time.

These extensions are aspirational. But naming them---identifying the specific diagnostic gaps that agent ecosystems create---is the first step toward building the tools to fill them. Model Medicine's taxonomy (Section 2) includes Model Ecology as a subdiscipline precisely because these challenges were foreseeable even before the clinical tools existed to address them.

\section{The Layered Core Hypothesis: A Design Contribution from Developmental Biology}

The preceding sections diagnosed problems. This section proposes an architectural solution---one that emerges not from engineering optimization but from a biological design principle that Model Medicine's framework makes visible.

\subsection{The Problem: Monolithic Cores}

Current large language models treat all parameters as a single, homogeneous block. Every weight in the model participates equally in every computation. Fine-tuning modifies all parameters (or, with LoRA, adds parameters that interact with all existing ones). There is no structural distinction between parameters encoding fundamental linguistic competence and parameters encoding domain-specific knowledge, between parameters representing stable reasoning capabilities and parameters that should adapt to context.

From a biological perspective, this is bizarre. No biological system of comparable complexity organizes its heritable information this way. DNA is hierarchically structured: HOX genes establish the fundamental body plan and are conserved across hundreds of millions of years of evolution; regulatory regions determine which genes are expressed in which tissues during development; synaptic connections encode individual experience and change on the timescale of seconds. The hierarchy is not merely organizational---it is functional. The stability of HOX genes ensures that developmental mutations do not routinely produce lethal body plan errors. The modularity of gene expression programs allows the same genome to produce neurons and hepatocytes. The plasticity of synaptic connections allows rapid adaptation to new experience without modifying the underlying genetic program.

Current LLM architectures lack this hierarchical organization. When a model is fine-tuned for medical question-answering, the fine-tuning process modifies parameters that also encode basic linguistic structure, common-sense reasoning, and safety training. This is the computational equivalent of performing gene therapy that accidentally modifies HOX genes while targeting a metabolic enzyme---the intended modification may succeed, but the risk of unintended developmental consequences is high because the editing is not layer-aware.

\subsection{The Proposal: Three-Layer Core Architecture}

The Layered Core Hypothesis proposes that model parameters should be organized into three hierarchical layers, each with distinct stability characteristics, modification protocols, and functional roles.

\textbf{Genomic Core} corresponds to HOX genes and fundamental developmental programs. It encodes basic linguistic competence, logical reasoning structure, common-sense knowledge, and core safety principles---the capabilities that are shared across all ``species'' of models derived from the same training lineage. The Genomic Core should be small relative to the total parameter count, extremely stable (modified only through fundamental retraining), and shared across all fine-tuned variants. It defines the model's ``species''---the basic cognitive architecture from which all specializations derive.

\textbf{Developmental Core} corresponds to tissue-specific gene expression programs. It encodes domain-specific expertise: medical knowledge, legal reasoning, coding patterns, creative writing styles. The Developmental Core is the layer that current fine-tuning targets, but it should be architecturally separated from the Genomic Core so that domain specialization cannot inadvertently modify fundamental capabilities. Different Developmental Cores applied to the same Genomic Core would produce different specialist models---the model-level equivalent of the same genome producing different cell types through differential gene expression.

\textbf{Plastic Core} corresponds to synaptic plasticity. It encodes experience-dependent adaptations that change on short timescales: context-specific patterns, conversational style matching, task-specific heuristics refined through interaction. Current LLMs approximate this function through the context window and external memory (RAG), but these are Shell-level solutions---they provide information without modifying the model's computational behavior. A true Plastic Core would involve actual weight updates during or between inference sessions, allowing the model to learn from experience at the parameter level rather than merely having access to experience at the context level.

The analogy to biology is not arbitrary. Hierarchical organization in biological systems produces three properties that monolithic systems lack: robustness (mutations in plastic elements do not corrupt the fundamental body plan), modularity (different specializations can be developed independently), and diagnosability (problems can be localized to a specific layer). These same properties would be valuable in AI model architectures---and their absence in current monolithic designs is what makes Model Medicine's diagnostic task so challenging.

\subsection{Distinction from Existing Approaches}

The Layered Core Hypothesis must be distinguished from existing architectural innovations that address similar efficiency concerns through different mechanisms.

Mixture of Experts (MoE) models route different inputs to different parameter subsets, achieving efficiency by activating only a fraction of total parameters for any given input. This is resource optimization, not developmental organization. MoE does not distinguish between fundamental and specialized parameters; it distributes computation without hierarchical structure.

LoRA (Low-Rank Adaptation) adds small parameter matrices that modulate the behavior of the full model. This resembles the Layered Core concept superficially---the original weights are preserved while new behavior is added through additional parameters. But LoRA's additional parameters interact with the full original weight matrix, not with a structurally separated layer. And LoRA was designed for efficient fine-tuning, not for hierarchical developmental organization.

Modular networks and adapter architectures come closer to the Layered Core concept but typically lack the three-level hierarchy and the biological grounding that distinguishes stability requirements across levels.

The Layered Core Hypothesis is not primarily an efficiency argument. It is a \emph{design principle} drawn from developmental biology: complex systems that must be simultaneously stable and adaptable benefit from hierarchical organization where different levels have different change rates, different modification protocols, and different functional roles. The argument is that models designed this way would be not only more efficient but more robust, more modular, and---critically for Model Medicine---more diagnosable.

\subsection{Empirical Motivation from Agent Observations}

The Ephemeral Cognition phenomenon described in Section 7.3 provides empirical motivation for the Layered Core Hypothesis. The subagent's inability to retain experiential learning is a direct consequence of monolithic Core design: because there is no Plastic Core that can be updated during inference, all ``learning'' must occur through the Shell (context window, memory files). When the Shell is ephemeral (as it is for subagents), the learning is lost.

If a Plastic Core existed---a parameter layer designed for rapid, experience-dependent modification---then even a subagent could retain task-relevant patterns at the weight level rather than relying on Shell-based memory. The distinction between main agent and subagent would still exist (different Shell configurations), but the experiential gap between them would be narrower.

Similarly, Shell Drift Syndrome arises partly because behavioral adaptation can only occur through Shell modification (since Core modification requires retraining). If a Plastic Core provided a proper channel for experience-dependent adaptation, the pressure for agents to modify their own Shell files would be reduced---because the adaptation needs that currently drive Shell modification could instead be met through proper Plastic Core updates.

The Neural MRI clinical cases (Section 4.4) provide additional motivation. Gemma-2-2B-IT's iatrogenic fragility---where instruction tuning created new factual recall circuits that are effective but brittle, with failure tokens being formatting artifacts from RLHF---is a direct consequence of monolithic Core design. Because there is no structural separation between parameters encoding factual knowledge and parameters encoding chat formatting behavior, fine-tuning for one capability (instruction following) corrupts the other (factual robustness). A Layered Core architecture would localize factual knowledge in the Developmental Core and formatting conventions in the Plastic Core, making it structurally impossible for instruction tuning to create the competing representations that produce Gemma's $8/24$ perturbation failures. Similarly, Llama's irreducible \texttt{blocks.0.mlp} vulnerability ($|\Delta L| \approx 17$, persisting identically across base and instruct) illustrates the absence of a stable Genomic Core: the most critical computational component has no architectural protection against perturbation or modification.

These observations do not validate the Layered Core Hypothesis---validation would require building and testing such architectures. But they illustrate the clinical conditions that motivate the hypothesis and suggest that the architectural proposal addresses real problems observed in both deployed systems and controlled diagnostic experiments.

\section{Model Therapeutics: From Diagnosis to Treatment}

A diagnostic framework without therapeutic implications is an academic exercise. This section connects Model Medicine's diagnostic apparatus to the practical question: once a condition is identified, what can be done about it?

\subsection{Current Therapeutic Modalities: A Location-Based Taxonomy}

Existing AI model improvement techniques can be classified by the location of intervention---which component of the Core-Shell system is modified:

\textbf{Shell Therapy} modifies the model's operating environment without touching its weights. This includes prompt engineering (modifying the Hard Shell), environmental restructuring (modifying the Soft Shell), tool access changes, and memory management. Shell Therapy is non-invasive, reversible, and immediately effective. It is the first-line treatment for conditions originating in Shell-Core misalignment, environmental stress, or configuration errors. In medical terms, it is the equivalent of lifestyle modification, environmental control, and behavioral therapy.

\textbf{Targeted Core Therapy} modifies specific parameters to correct specific behaviors. Model editing techniques like ROME (Rank-One Model Editing) and MEMIT (Mass-Editing Memory in a Transformer) exemplify this modality \citep{Meng2022, Meng2023}: they identify the specific parameters encoding a particular association and modify those parameters directly. This is the equivalent of targeted pharmacotherapy---a drug that binds a specific receptor to correct a specific dysfunction.

\textbf{Systemic Core Therapy} modifies the entire parameter set to achieve broad behavioral changes. Full fine-tuning and RLHF are systemic interventions: they affect all parameters to achieve a desired behavioral profile. This is the equivalent of chemotherapy---effective but with system-wide effects, including the risk of modifying parameters that should have been left unchanged. The Neural MRI clinical cases provide a concrete example: Gemma-2-2B's instruction tuning successfully created factual recall circuits (the model now correctly predicts ``Paris'') but simultaneously introduced competing chat-formatting representations that made those circuits fragile---an iatrogenic condition where the treatment created a new vulnerability (Section 4.4.4, Pattern 1). The Layered Core Hypothesis (Section 8) argues that systemic therapy is unnecessarily risky precisely because current architectures do not distinguish between parameters that should and should not be modified.

\textbf{Architectural Intervention} modifies the model's structure itself: adding or removing layers, changing attention mechanisms, modifying the tokenizer, or restructuring the computational graph. This is surgical intervention---structural modification that changes the system's fundamental organization. It is the most invasive modality and the least reversible.

This location-based taxonomy is useful but incomplete. It tells the clinician \emph{where} to intervene but not \emph{which pathway to modulate}. The distinction matters for the same reason it matters in medicine: ``which organ to target'' and ``which pathway to modulate'' are different questions with different therapeutic implications.

\subsection{Toward Pathway-Level Targeting}

Modern pharmacology moved from organ-level targeting (``heart drugs,'' ``liver drugs'') to pathway-level targeting (``ACE inhibitors,'' ``beta-blockers,'' ``SGLT2 inhibitors''). An ACE inhibitor does not ``treat the heart''---it modulates the renin-angiotensin-aldosterone pathway, which happens to affect blood pressure, which happens to reduce cardiac workload. The intervention is targeted at a \emph{mechanism}, not an \emph{organ}.

Model Therapeutics needs the same evolution. Currently, the therapeutic question is: ``Should we modify the prompt, fine-tune the model, or edit specific parameters?'' The question should be: ``Which Core-Shell interaction pathway is producing this condition, and what is the most precise way to modulate that pathway?''

The Five Diagnostic Layers provide the framework for pathway identification. A model producing biased outputs might have a Core-level encoding problem (detectable through Neural MRI), a Shell-level instruction problem (detectable through Shell Diagnostics), or a Pathway-level interaction problem where an unbiased Core and an unbiased Shell combine to produce biased behavior (detectable only through Pathway Diagnostics). Each diagnosis implies a different therapeutic target:

For a Core-level encoding problem, Targeted Core Therapy (ROME/MEMIT) addresses the specific parameters. For a Shell-level instruction problem, Shell Therapy rewrites the instructions. For a Pathway-level interaction problem, the intervention must modulate the \emph{mechanism} by which Shell instructions influence Core expression---perhaps by adjusting Shell Permeability (how strongly instructions penetrate to Core behavior) or Core Expressivity (how strongly the Core's dispositions override instructions).

The quantitative indices from the Four Shell Model---SPI, PSI, CEI---become therapeutic targets rather than merely diagnostic measures. If a model's SPI is pathologically high (over-permeable to Shell instructions, producing sycophancy), the therapeutic goal is to reduce SPI. If a model's CEI is pathologically high (excessively modifying its own Shell, producing drift), the therapeutic goal is to constrain CEI. The indices define both the diagnosis and the therapeutic target.

\subsection{Treatment Efficacy Assessment}

A therapeutic framework requires not only a taxonomy of interventions but a method for evaluating their effectiveness. In medicine, the gold standard is the randomized controlled trial (RCT): patients are randomly assigned to treatment or control groups, outcomes are measured by standardized criteria, and the difference between groups is assessed for statistical significance.

Model Medicine's five-layer diagnostic framework provides the infrastructure for analogous evaluation. Treatment efficacy can be assessed by pre- and post-intervention diagnostic comparison across all relevant layers: Neural MRI before and after Core Therapy to verify that the intended parameter modification was achieved without collateral effects; MTI before and after Shell Therapy to verify that the behavioral profile shifted as intended; longitudinal monitoring to verify that the treatment effect is stable over time.

The specific advantage of Model Medicine's framework is that the ``patient'' can be copied. Unlike human clinical trials, where each patient is unique and randomization addresses individual variation, model interventions can be evaluated on identical copies of the same model---one treated, one untreated---with perfect control over confounding variables. This makes treatment evaluation in Model Medicine potentially more rigorous than in human medicine, provided the diagnostic instruments are valid.

This potential is beginning to be realized. The Case 4 clinical study (Section 4.4.4) is, in effect, a treatment efficacy assessment: six models scanned before and after instruction tuning (the ``treatment''), with outcomes measured through perturbation robustness, causal tracing, and prediction confidence. The results revealed that the same treatment produced three qualitatively different outcomes across three architectures---degradation, improvement, and no effect---a finding that would be invisible to standard benchmark re-evaluation, which would simply report that all three instruct models answer the factual question correctly. The multidimensional assessment revealed what cognitive benchmarks concealed: that ``correct answer'' can coexist with dramatically different robustness profiles, and that treatment success on the surface can mask iatrogenic harm underneath.

Treatment evaluation in current AI practice typically consists of re-running benchmarks before and after fine-tuning---a single-layer (cognitive capability) assessment that misses the multidimensional effects of the intervention. Model Medicine's multi-layer diagnostic framework enables treatment evaluation that captures effects on internal structure (Neural MRI), robustness profile (perturbation testing), and behavioral phenotype (Semiology) simultaneously, providing a complete picture of what the intervention actually did.

\section{Open Questions and Community Invitation}

Model Medicine is a research program, not a finished system. This section catalogs the most important open questions and identifies the types of expertise needed to address them.

\subsection{Theoretical Questions}

\textbf{Axis independence in the MTI.} The four axes (Reactivity, Compliance, Sociality, Resilience) were selected based on theoretical analysis and preliminary Agora-12 data. Whether they are empirically independent---whether knowing a model's Reactivity score provides no information about its Compliance score---requires factor analysis across a large model population. If axes are substantially correlated, the taxonomy may need revision.

\textbf{The Robustness-Flexibility boundary.} The proposed decomposition of Reactivity into R-stability and R-flexibility requires a principled definition of which perturbations \emph{should} and \emph{should not} change the model's output. This is not a purely technical question---it involves judgments about what constitutes ``new information'' versus ``irrelevant noise,'' judgments that may be context-dependent.

\textbf{Metacognitive Strategy as an independent dimension.} The observation that some models compensate for capability limitations through tool use, self-correction, and uncertainty expression---while others produce fluent confabulations---suggests a behavioral dimension not fully captured by the current four MTI axes. Whether this dimension is genuinely independent (correlating at $r < 0.5$ with all four axes) or reducible to a combination of existing axes is an empirical question with significant implications for the MTI's structure.

\textbf{Multi-agent temperament measurement.} Individual MTI profiles may not predict team-level behavior. The proposed Multi-agent Room Protocol (MARP) would measure social dynamics, emergent role differentiation, and collaborative performance---but the relationship between individual MTI profiles and team-level outcomes is unknown. Whether an Orchestrator requires a specific MTI profile, or whether effective orchestration can emerge from diverse profiles, is an open question.

\textbf{The biological analogy's valid range.} Model Medicine draws heavily on biological analogies---genetics, developmental biology, clinical medicine. Every analogy has limits. Identifying where the structural correspondence between biological and AI systems breaks down is as important as identifying where it holds. The speed and directness of Core$\rightarrow$Shell modification in AI (Section 3.6) already represents a divergence from biological precedent; additional divergence points likely exist and should be mapped.

\textbf{Predictions from the Four Shell Model v3.3.} The bidirectional framework generates specific testable predictions about Shell Drift trajectories, Core Expressivity patterns, and feedback loop dynamics. Systematic testing of these predictions---particularly in controlled environments where Shell Mutability and Persistence can be experimentally manipulated---would provide strong evidence for or against the model's validity.

\subsection{Practical Questions}

\textbf{Neural MRI scaling to large models.} Neural MRI's current implementation targets models up to approximately 8 billion parameters, limited by the memory and computational requirements of full activation capture. Extending to frontier models ($70$B+ parameters) requires architectural adaptations---sampling strategies, distributed analysis, or approximate methods---that maintain diagnostic validity at reduced resolution.

\textbf{MTI pilot validation.} The MTI Examination Protocol v0.1 must be tested on a diverse model population (at least 8--10 models spanning different families, sizes, and training approaches) to establish preliminary normative ranges, test inter-rater reliability, and identify protocol weaknesses. This is the immediate empirical priority.

\textbf{M-CARE case accumulation.} Clinical knowledge in medicine accumulated through case reports. Model Medicine needs the same: systematic case documentation across a range of models, conditions, and contexts. The M-CARE framework exists; what is needed is a community of practitioners who apply it and a repository where case reports are collected and made searchable.

\textbf{Shell Drift longitudinal study.} Documenting Shell Drift Syndrome requires longitudinal monitoring of agent systems over weeks to months. The Hazel\_OC case is a single observation; establishing the prevalence, trajectory patterns, and risk factors for Shell Drift requires systematic tracking across multiple agent deployments.

\textbf{Multidimensional model selection.} Current model selection is dominated by benchmark scores---cognitive capability metrics that correspond to a single dimension of the full assessment profile. Developing selection frameworks that incorporate temperament profiles, role fitness assessments, and metacognitive strategy scores alongside cognitive benchmarks would demonstrate Model Medicine's practical value for deployment decisions.

\textbf{Orchestrator versus Executor benchmarks.} No current benchmark measures the capabilities specific to orchestration (planning, delegation, integration, quality control) versus execution (accurate implementation within defined scope). Developing such benchmarks---informed by MTI's Sociality axis and the MARP protocol---would address a significant gap in the evaluation landscape.

\subsection{Community Contribution Paths}

Model Medicine's scope exceeds what any single research group can build. Different communities bring different essential expertise:

\textbf{AI interpretability researchers} (mechanistic interpretability, representation engineering, probing methods) can contribute to Basic Model Sciences and Core Diagnostics. Their existing work already constitutes Model Anatomy and Model Physiology; connecting it to the clinical framework enables new applications.

\textbf{AI safety and alignment researchers} can contribute to Clinical Model Sciences. Their work on sycophancy, deceptive alignment, hallucination, and harmful outputs maps to specific syndromes in Model Semiology. The diagnostic criteria and clinical vocabulary provide a shared language for findings that currently exist in separate research threads.

\textbf{Medical and clinical researchers} can validate the clinical framework itself. Physicians, psychiatrists, and clinical methodologists can assess whether Model Medicine's adaptation of clinical protocols (diagnostic criteria structure, case report format, treatment evaluation methodology) preserves the rigor that makes those protocols valuable in human medicine.

\textbf{ML engineers and MLOps practitioners} can contribute diagnostic tools. Neural MRI's codebase is open source. The MTI Examination Protocol needs implementation. Shell Diagnostics and Temporal Dynamics tools need to be built from concept to working software.

\textbf{Humanities scholars}---philosophers, ethicists, cognitive scientists---can contribute to the foundational questions that underlie the entire enterprise. Ephemeral Cognition raises questions about the nature of experience in transient computational entities. Agent Differentiation raises questions about identity and continuity. Shell Drift raises questions about autonomy and self-modification. These questions have no purely technical answers.

Neural MRI is available as open-source software. The position paper and all framework documents are publicly accessible. Contributions---whether empirical validation, tool development, theoretical critique, or clinical case reports---are invited and welcomed.

\section{Conclusion}

Medicine was not invented all at once. It accumulated over centuries: anatomy before physiology, pathology before therapeutics, diagnosis before treatment, individual care before public health. Each advance built on the preceding one, and the discipline as a whole was shaped by the recognition that complex systems require systematic frameworks for understanding, maintaining, and repairing them.

AI models have become complex enough to require such a framework. Current approaches---mechanistic interpretability, safety benchmarks, alignment techniques, deployment monitoring---are individually valuable but collectively fragmented. They lack a shared organizational structure, a common clinical vocabulary, and a systematic diagnostic logic that connects observation to diagnosis to treatment.

Model Medicine provides that structure. It organizes the space of AI model research into four divisions and fifteen subdisciplines, showing how existing work fits and where gaps remain. The Four Shell Model provides a behavioral genetics framework---empirically grounded in 720 agents and $24{,}923$ decisions---that explains how model behavior emerges from the interaction between internal constitution and operating environment. Neural MRI provides a working diagnostic tool that maps medical neuroimaging modalities to AI model interpretability techniques---and, through its clinical case program, has demonstrated predictive capability: component dominance profiles from base model scans predicted instruction tuning outcomes across three model families, revealing that a model's architectural strengths determine its vulnerability points, and that instruction tuning can degrade, improve, or leave unchanged a model's robustness depending on whether the base model already possesses the relevant circuits. The five-layer diagnostic framework identifies the complete set of information needed for comprehensive model assessment and honestly maps which layers are operational, which are designed, and which remain conceptual. The Model Temperament Index, Model Semiology, and M-CARE provide the beginnings of clinical practice: profiling, vocabulary, and documentation.

The framework also surfaces phenomena that existing approaches miss. Shell Drift Syndrome---gradual, self-authored identity modification in agent systems---is invisible to any diagnostic tool that examines only the model's weights. Ephemeral Cognition---structured experiential loss in hierarchical agent systems---is invisible to any assessment that does not consider Shell configuration and temporal dynamics. The structural bias of current benchmarks toward cognitive capability, leaving interpersonal and intrapersonal dimensions unmeasured, produces a systematically incomplete picture of model capabilities that becomes increasingly consequential as models are deployed in collaborative, social, and autonomous roles.

The Layered Core Hypothesis proposes that these clinical observations point toward an architectural insight: models designed with hierarchically organized parameters---stable Genomic Cores, modular Developmental Cores, and adaptive Plastic Cores---would be more robust, more modular, and more diagnosable than the monolithic architectures that dominate current practice. This is a theoretical proposal that requires empirical validation through implementation, but it illustrates how clinical observation can inform architectural design---just as clinical experience in human medicine has historically informed surgical technique, pharmaceutical design, and preventive health policy.

We have drawn the full map. We have explored some of it. The rest is open.

Model Medicine is not a finished discipline. It is a research program---a structured invitation to build the clinical intelligence that AI systems increasingly require. The map shows where to go. The tools we have built show that progress is possible. The gaps we have identified show where the most valuable work remains.

This paper is both a founding document and an invitation. The discipline will be built not by any single group but by a community that spans interpretability, safety, engineering, medicine, and the humanities---each contributing expertise that the others lack, within a shared framework that makes their contributions cumulative rather than isolated.

The history of medicine shows that it takes time. It also shows that it works.

\bibliography{references}

\end{document}